\documentclass[10pt,twocolumn,letterpaper]{article}

\usepackage[pagenumbers]{cvpr} 

\usepackage{graphicx}
\usepackage{amsmath}
\usepackage{amssymb}
\usepackage{booktabs}
\usepackage{enumitem}

\usepackage[pagebackref,breaklinks,colorlinks]{hyperref}

\usepackage[super]{nth}
\usepackage{bm}
\usepackage{makecell}
\usepackage{color, colortbl}
\usepackage{multirow}
\usepackage{amssymb}
\usepackage{pifont}
\usepackage{caption}
\usepackage{siunitx} 
\usepackage{arydshln} 

\newcommand\mypara[1]{\vspace{1.0mm}\noindent\textbf{#1}}
\DeclareMathOperator*{\argmax}{argmax}
\newcommand\nl[1]{{\it``#1''}} 
\newcommand\fl[1]{{\fontfamily{phv}\selectfont\footnotesize#1}}

\newcommand{\milestone}{\textsc{M-Track}\xspace}
\newcommand{\Milestone}{\textsc{M-Track}\xspace}
\newcommand{\milestonelong}{milestone-based task tracker\xspace}

\newcommand{\Milestonelongcap}{Milestone-based Task Tracker\xspace}

\newcommand{\BaseLSTM}{LSTM\xspace}
\newcommand{\EnhancedLSTM}{LSTM-L\xspace}
\newcommand{\BaseVLNBERT}{VLN$\circlearrowright$BERT\xspace}
\newcommand{\EnhancedVLNBERT}{VLN$\circlearrowright$BERT-L\xspace}

\newcommand{\LSTM}{\EnhancedLSTM}
\newcommand{\VLNBERT}{\EnhancedVLNBERT}

\newcommand{\LSTMMilestone}{\EnhancedLSTM + \milestone\xspace}
\newcommand{\VLNBERTMilestone}{\EnhancedVLNBERT + \milestone\xspace}

\newcommand{\nop}[1]{}
\newcommand{\cmark}{\ding{51}}

\definecolor{Gray}{gray}{0.93}

\usepackage[capitalize]{cleveref}
\crefname{section}{Sec.}{Secs.}
\Crefname{section}{Section}{Sections}
\Crefname{table}{Table}{Tables}
\crefname{table}{Tab.}{Tabs.}

\begin{document}

\title{One Step at a Time:\\ Long-Horizon Vision-and-Language Navigation with Milestones}

\author{Chan Hee Song$^1$
\and
Jihyung Kil$^1$
\and
Tai-Yu Pan$^1$
\and
Brian M. Sadler$^2$
\and
Wei-Lun Chao$^1$
\and
Yu Su$^1$ \\
$^1$The Ohio State University \quad $^2$U.S.\ Army Research Laboratory \\
\texttt{\small \{song.1855, kil.5, pan.667, chao.209, su.809\}@osu.edu} \\
\texttt{\small brian.m.sadler6.civ@army.mil}
}

\maketitle


\begin{abstract}

    We study the problem of developing autonomous agents that can follow human instructions to infer and perform a sequence of actions to complete the underlying task.
    Significant progress has been made in recent years, especially for tasks with short horizons. However, when it comes to long-horizon tasks with extended sequences of actions, an agent can easily ignore some instructions or get stuck in the middle of the long instructions and eventually fail the task. 
    To address this challenge, we propose a model-agnostic \milestonelong (\milestone) to guide the agent and monitor its progress. 
    Specifically, we propose a milestone builder that tags the instructions with navigation and interaction milestones which the agent needs to complete step by step, and a milestone checker that systemically checks the agent's progress in its current milestone and determines when to proceed to the next.
    On the challenging ALFRED dataset, our \milestone leads to a notable $33\%$ and $52\%$ relative improvement in unseen success rate over two competitive base models. Check our code at \url{https://github.com/chanhee-luke/M-Track}.
\end{abstract}


\section{Introduction}
\label{s_intro}

As autonomous agents (\eg, robots) become more integrated into our daily life, it is increasingly important to develop autonomous agents that can understand natural language commands and carry out the corresponding tasks. To facilitate such a goal, various benchmarks have been proposed in the realm of robot instruction following such as vision-and-language navigation (VLN)~\cite{anderson2018vision,chen2019touchdown,regneri2013grounding,das2018embodied,wijmans2019embodied,jain2019stay,gordon2018iqa,puig2018virtualhome,zhu2017visual}, together with a number of novel algorithms that consistently push forward the state of the art~\cite{ma2019self, tan2019learning, Ma_2019_CVPR, wang2020active}. 
Specifically, to succeed in VLN, an agent must comprehend the language instruction, ground it into the partially-observable environment with only visual perception, and plan and perform navigation and interaction actions in the environment to complete the task.

\begin{figure}
    \centering
    \includegraphics[width=0.97\linewidth]{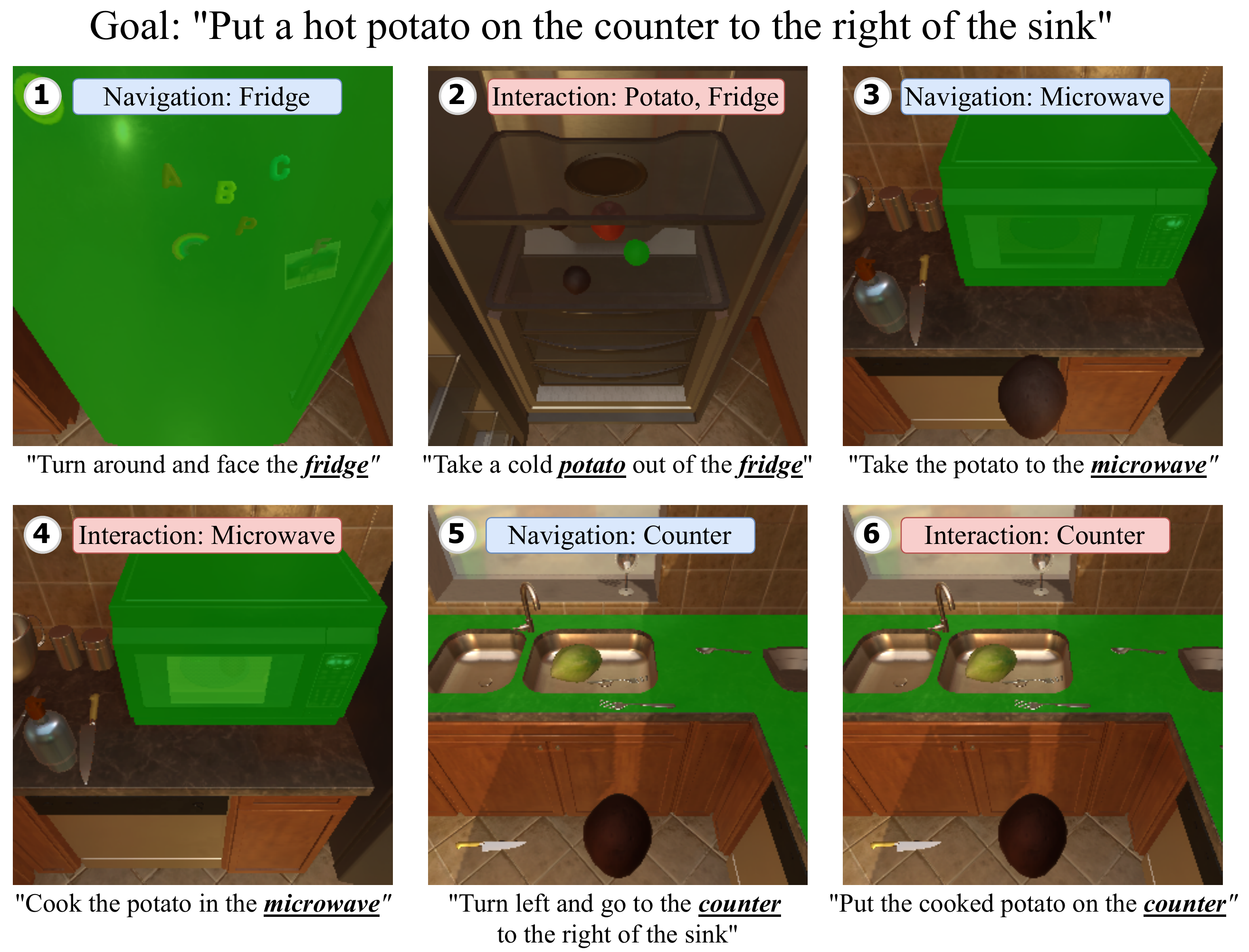}
    \vskip -8pt
    \caption{\textbf{Illustration of our \milestone approach}. We show an ALFRED task~\cite{ALFRED20}, which consists of an overall goal (text on the top) and six subtasks (text below each image). The {\color{cyan}blue}/{\color{magenta}red} text box within each image is our extracted {\color{cyan}navigation}/{\color{magenta}interaction} milestones from the subtask instructions. An agent needs to reach the milestone of the current subtask (\eg, reaching proximity to the target object for navigation milestones, or having interacted with the target objects for interaction milestone; {\color{green}green} masks for target objects) before it can proceed to the next subtask.}
    \vskip -10pt
    \label{fig_alfred}
\end{figure}

One critical challenge in VLN arises when the task horizon becomes substantially longer~\cite{ALFRED20}. That is, a task is so complex that it essentially consists of multiple 
``subtasks'' that need to be completed \emph{sequentially} to fulfill the whole task.
For example, in \autoref{fig_alfred} the task \nl{put a hot potato on the counter to the right of the sink} can be decomposed into six subtasks. Moreover, the subtask \nl{heat the potato} must be carried out before the subtask \nl{put the potato on the counter}; otherwise, the final task is doomed to fail no matter how accurate the subsequent planning is. Such a sequential dependency requires the agent to closely monitor its progress and ensure it is staying on the right track when carrying out a long-horizon task.

At first glance, this challenge may seem trivial if the language instruction is detailed enough (like in \autoref{fig_alfred}), such that it already defines the subtasks and their order.
However, as shown in the literature~\cite{singh2021factorizing,wang2020active, ke2019tactical, ma2019self, zhang2021hierarchical, blukis2021persistent} and our experiments, an agent fed with detailed instructions still frequently skips subtasks, or wanders around within a subtask even when it is already completed. In essence, what an agent truly struggles with is the lack of awareness of \emph{where it currently is in the long subtask sequence and how much progress it has made within a subtask}.

To address this issue, we propose to equip VLN agents with an \emph{explicit task tracker}, which keeps track of the agent's progress within a subtask and guides it for when to move on to the next. Concretely, we propose the concept of \textit{milestone}, which renders the necessary condition of completing a subtask. Namely, for a subtask to be considered as completed, the milestone must be reached.
Take the subtask \nl{take a cold potato out of the fridge} in \autoref{fig_alfred} as example. To complete it, the necessary condition is that the agent must see the potato and the fridge, be close enough to them, and perform an interaction action with the potato. 
We argue that by explicitly extracting such milestones from the instructions and \emph{grounding} them to the environment state, we can systematically determine if the agent should continue working on the current subtask or proceed to the next. 

To this end, we propose the \textit{\milestonelong} (\milestone), which consists of two components: \textit{milestone builder} and \textit{milestone checker}. The milestone builder extracts the milestone (\ie, the necessary completion condition) of each subtask from the corresponding language instruction. We model it as a named entity recognition problem and train a BERT-CRF tagger~\cite{devlin2018bert, Souza2019PortugueseNE} to accurately extract both the target objects and their action type (\ie, navigation or interaction).
The milestone checker then tries to \emph{ground} (\ie, identify and localize) the extracted target objects in the perceived environment using an object detection model~\cite{he2017mask} and 
checks if the agent is close enough to them and/or is about to interact with them --- to decide if the agent is completing the current subtask and ready to move on. It is worth noting that our \milestone only needs to access the language instructions, the visual input to the agent, and the agent's action, not any internal states of the agent. Thus, it is \textit{model-agnostic} and can be easily integrated with any agent model with minimal changes.
 
\begin{figure*}[!th]
    \centering
    \includegraphics[width=\linewidth]{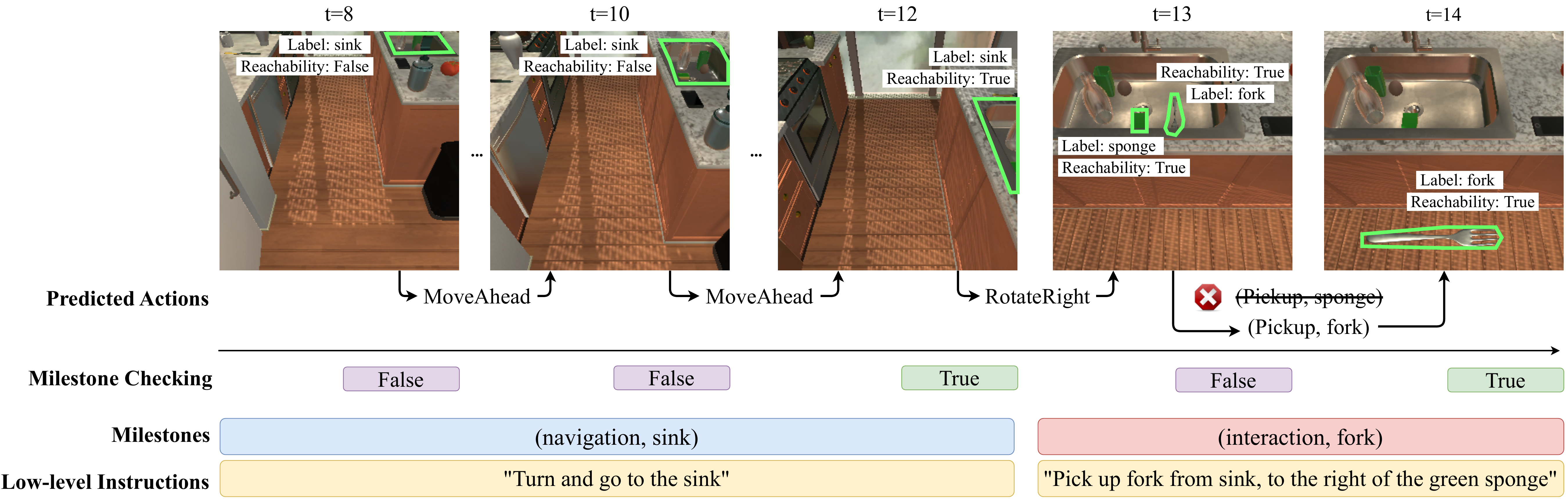}
    \vskip -5pt
    \caption{\textbf{Overview of the milestone checking process}. Milestones are extracted from the current low-level instruction by our milestone builder (\S\ref{sec:milestone_builder}). After an action is predicted, our milestone checker (\S\ref{milestone_checking}) examines, based on objects with reachability information (text in images) from its object detector, if the resultant state satisfies the milestone. Only when the milestone is satisfied, the next low-level instruction is provided to the agent. The agent is prevented from picking up a wrong object (sponge) by our proactive checking (\S\ref{sec:planning_with_milestone}).} 
    \vskip -10pt
    \label{fig_milestone_checking}
\end{figure*}

\emph{How can \milestone interacts with the agent to affect its action (\eg, to not skip a subtask)?} We propose two simple yet effective ways. First, at any time step, we feed the agent with only the part of the instructions that corresponds to the current subtask determined by the milestone tracker. This explicitly guides the agent to focus on the current subtask. Second, and more importantly, we apply the milestone checker \emph{proactively} --- before the agent executes its predicted action --- to reject actions that will lead to subtask failures. For instance, we reject the action of taking a \nl{sponge} if the milestone object is \nl{fork} (\autoref{fig_milestone_checking}).

We validate \milestone on ALFRED \cite{ALFRED20}, a recently released large-scale VLN dataset for common household tasks. The tasks in ALFRED are considered long-horizon because on average each task needs \num{50} actions to complete. In contrast, another popular dataset R2R \cite{anderson2018vision} needs only \num{5}. 
We integrate \milestone into two baseline VLN models \BaseLSTM~\cite{ALFRED20} and \BaseVLNBERT~\cite{Hong_2021_CVPR}, and demonstrate notable and consistent performance gains. When tested in seen environments, \milestone leads to $16\%$--$57\%$ relative improvement in success rate. In more challenging unseen environments, the relative gain increases to $33\%$--$52\%$. Our ablation studies and qualitative results further verify that the improvement indeed comes from agents able to better follow the sequence of subtasks and stay on the right track.


\section{Related Work}
\label{s_related}

\mypara{VLN datasets.}
Significant efforts have been devoted to creating simulated environments and datasets for VLN, where a virtual agent has egocentric perception of the environment and takes actions to navigate in it~\cite{anderson2018vision,chen2019touchdown,regneri2013grounding,das2018embodied,wijmans2019embodied,jain2019stay,gordon2018iqa,puig2018virtualhome,zhu2017visual}. However, most datasets do not consider interaction actions with objects, significantly limiting the complexity of tasks that an agent can perform. The recent ALFRED dataset~\cite{ALFRED20} is among the first to provide tasks that involve both navigation and interaction actions, providing a more challenging benchmark with much longer task horizons.

\mypara{VLN models.}
Most early VLN models follow an LSTM-based sequence-to-sequence architecture, taking language and visual sequences as input and predicting a sequence of actions~\cite{ALFRED20,anderson2018vision,fried2018speaker,ma2019self,tan2019learning,kurita2020generative}. Because of the recent success of the Transformer~\cite{vaswani2017attention} in vision tasks, Transformer-based models are increasingly adopted for VLN~\cite{suglia2021embodied,pashevich2021episodic,li2019robust,majumdar2020improving,Hong_2021_CVPR}. Our \Milestone is model-agnostic and is compatible with models of both types (cf.\ \S\ref{s_Model}).

\mypara{Natural language instructions.} ALFRED provides each task with both a high-level (\ie, goal) instruction and more detailed low-level instructions. Most previous studies train the agent with the whole instruction (\ie, concatenation of the high-level and low-level instructions) at each time step~\cite{ALFRED20,pashevich2021episodic,suglia2021embodied,singh2021factorizing,kim2021agent}. However, for long-horizon tasks like those in ALFRED, the low-level instructions can be quite long (six sentences on average). An agent fed with the whole instruction thus could have difficulty digesting the long instruction and easily lose track of the progress. \Milestone helps agents focus their attention on the most pertinent instruction and reduce distraction.

\mypara{Step-by-step language guidance.} 
To address the issues with long instructions, learning low-level instructions step by step has been explored in several prior studies~\cite{min2021film,zhu2020babywalk,zhang2021hierarchical,das2018neural,hong2020sub}.  BabyWalk~\cite{zhu2020babywalk} learns the low-level instruction step by step using curriculum learning. HiTUT~\cite{zhang2021hierarchical} decomposes the whole instruction into hierarchical sub-problems and learns sequentially with a hierarchical task network. Concurrent to this work, FILM~\cite{min2021film} decomposes the instruction into subtasks and learns them sequentially with the help of a semantic map.
\Milestone shares a similar rationale.
However, \milestone is notably different from existing methods, especially regarding when to feed the next low-level instruction during \emph{test time}. 
First, \milestone explicitly and systematically checks the agent's progress, during both training and test time, by 1) defining the completion condition, \ie, the milestone, of each subtask and 2) verifying the milestone by grounding it into the environment via a visual object detector. In contrast, existing methods either train a binary classifier to determine subtask completion~\cite{das2018embodied}, or simply set an upper bound for the number of actions to execute within each subtask~\cite{zhang2021hierarchical,zhu2020babywalk}, or only checks if the agent needs to stop using a separate module~\cite{xiang-etal-2020-learning}.  As will be seen in \S\ref{s_exp}, \milestone notably outperforms these methods in tracking the agent's progress and feeding the right instruction.
Second, \milestone also proactively guides the agent for better action prediction, creating another gain in performance (cf.\ \S\ref{sec:planning_with_milestone}). Finally, \milestone is not embedded in any specific VLN model; it is model-agnostic and can be easily integrated into different VLN models (cf.\ \S\ref{s_Model}).


\section{VLN Background}
\label{s_background}

A VLN task is generally defined as follows: given a language instruction $I$, an agent needs to infer and perform a sequence of actions $\{a_0, a_1, \cdots, a_t, \cdots\}$ in the environment $E$ to complete the task. In datasets like ALFRED~\cite{ALFRED20}, the instruction $I$ is composed of a high-level instruction $I_H$ and a list of low-level instructions $I_L$, as exemplified in \autoref{fig_alfred}. A VLN task can thus be represented by a tuple $(I, E, G)$, in which $G$ is the goal test of the task.

For an agent to perform the task, it will be placed in the environment $E$ and have a certain pose at time step $t$, from which it can receive a visual input $v_t$. Based on $v_t$ and the instruction $I$, the agent then predicts an action $a_t$, which can either be a navigation one that changes the agent's pose (\eg, \fl{MoveAhead}) or an interaction one that interacts with the environment (\eg, \fl{PickupObject}). The agent also needs to predict a binary mask for the target object if it predicts an interaction action. Both types of actions can potentially change the visual input $v_{t+1}$ of the next time step. The agent will stop when it believes the task has been completed. The final state of the environment 
is then compared with the goal state $G$ to determine task completion.

Following ALFRED, we discretize an agent's action space into \num{5} navigation action (\fl{MoveAhead, RotateRight, RotateLeft, LookUp, and LookDown}), \num{7} interaction actions (\fl{PickupObject, PutObject, OpenObject, CloseObject, ToggleOnObject, ToggleOffObject}, and \fl{SliceObject}), and \num{1} stop action (\fl{Stop}).

\mypara{Agent model.} Without loss of generality, we define an agent model as $a_t=f(v_t, I_t, h_t)$, where $h_t$ is the memory from the previous time steps (\eg, the hidden state of an LSTM). $a_t$ is a tuple \fl{(action, object mask)}; the mask is \fl{null} for stop and navigation actions. $I_t$ is the instruction input at time $t$, which can be the entire $I$ or a portion of it.

\section{\Milestonelongcap (\milestone)}
\label{s_approach}

For long-horizon VLN tasks, an agent needs to complete multiple subtasks, usually in a specific order, to complete the whole task. 
More specifically, each low-level instruction in $I_L$ can be seen as a subtask.
Agents then have to decide, often implicitly, which subtask it is doing at each time step and when to move on to the next subtask, which itself is a challenging problem for the agent. 
To address that, we introduce an auxiliary module, \textit{\milestonelong} (\milestone), to explicitly and interactively guide the agent to make such decisions (see \autoref{fig_milestone_checking} for an overview). 

Next, we first introduce the design of \milestone (\S\ref{s_Milestone}), followed by how to integrate it with agent models (\S\ref{sec:planning_with_milestone}). We then introduce two base agent models (\S\ref{s_Model}) and how to train the base models with reinforcement learning (\S\ref{s_Learning}). 

\subsection{Design of \milestone}
\label{s_Milestone}
 
The core functionality of \milestone is to decide when an agent should move on to the next subtask. On the surface, this may be done simply by training a (binary) classifier, which takes all the language/visual signals as input. Doing so, however, does not exploit the fact that the (sub)tasks are compositional, composed of entities (\eg, objects) that are identifiable and localizable both in the environment and in the instruction. Leveraging the compositional nature of the (sub)tasks has multiple advantages. First, it reduces the input space for making the decision from the space of language/visual signals to that of discrete entities. Second, it makes the decision rule systematic and explainable: we can make the decision by directly comparing the entities detected in both modalities. Both of them could improve the generalizability of the decision function.  

We design \milestone to explicitly consider the compositional nature of (sub)tasks.
Specifically, we introduce the concept of \textit{milestone}, which is the \emph{necessary condition for completing a subtask}, \ie, an agent must reach the milestone in order for the corresponding subtask to be considered as completed. 
For example, if the subtask is \nl{move to the mug}, then the agent must \textbf{\color{red}navigate} to the \textbf{\color{blue}mug}, \textbf{see} it, and be \textbf{close enough} to the it. 
If the subtask is \nl{pick up the mug}, then the agent must \textbf{see} the \textbf{\color{blue}mug}, be \textbf{close enough} to it so that it can then  \textbf{\color{red}interact} with it. 
These two examples render the key ingredients of a milestone, which are its \textbf{\color{blue}target entities} and its \textbf{\color{red}type} (navigation or interaction). Meanwhile, we say an agent has reached a milestone only when it can perceive (see) the target entities, is already close to them, and is doing the right type of action with them.

To this end, we represent a milestone by a tuple \fl{({\color{red}type}, {\color{blue}target})}, and decompose our \milestone into two components: 1) a \textit{milestone builder} which constructs milestones from the low-level instructions $I_L$, and 2) a \textit{milestone checker} that checks if a milestone has been reached by an agent.

\subsubsection{Milestone Builder}
\label{sec:milestone_builder}

We generate the milestone of a subtask according to its corresponding low-level instruction in $I_L$ using named entity recognition~\cite{devlin2018bert}. For example, given an instruction \nl{Turn to the left and face the toilet}, the milestone builder should output the tag \fl{(navigation, toilet)}. For the instruction \nl{Pick the soap up from the back of the toilet}, the milestone builder should output \fl{(interaction, soap)}.  

For an interaction milestone, it should contain the target objects that the agent is going to newly interact with in the current subtask. 
For instance, if the subtask is \nl{Put down the potato on the counter} (\autoref{fig_alfred}), the agent is supposed to already be holding a potato (from previous subtasks). Thus, \nl{potato} is not a milestone target for the current subtask but \nl{counter} should be.
For a subtask that has multiple objects to be interacted with, the builder is designed to tag all of them. For instance, in the subtask \nl{Grab a potato from the fridge} (\autoref{fig_alfred}), the agent needs to 1) open the fridge, 2) pick up the potato, and 3) close the fridge. In this case, the builder tags both the potato and the fridge as the targets for an interaction milestone. In cases that the builder does not extract any target from the current subtask, it will merge the current subtask with the next one and use the milestone extracted from the next subtask.

Without loss of generality, we adopt a BERT-CRF model \cite{devlin2018bert, Souza2019PortugueseNE} for the milestone builder, and train it with data derived from the ALFRED training data. Training data is prepared using the metadata from the ALFRED simulator. More details are in the supplementary materials. We show that our milestone builder reaches a fairly high F1 score (see \autoref{tab_milestone_generation}). More analysis will be discussed in \S\ref{sec:component_analysis}.

\begin{table}
  \centering
  \small
  \renewcommand\arraystretch{0.8}
  \begin{tabular}{ccc}
    \toprule
    \textbf{Target Type} & \textbf{Val Seen} & \textbf{Val Unseen} \\
    \midrule
    Navigation & 90.16 & 90.62 \\
    Interaction & 96.85 & 97.17\\
    \bottomrule
  \end{tabular}
\caption{\small \textbf{F1 score of milestone builder on ALFRED validation.} 
}
  \label{tab_milestone_generation}
  \vskip -10pt
\end{table}

\subsubsection{Milestone Checker}
\label{milestone_checking}
 
We introduce a milestone checker that determines if an agent has reached a milestone (see \autoref{fig_milestone_checking}).
Specifically, we design it to be \emph{explicit}: we directly estimate the state of the agent/environment from the visual input and compare it with the milestone.
A navigation milestone is reached if the target object is detected in the visual input and located within a reachable distance to the agent ($1.5$ meters in ALFRED). An interaction milestone is reached with an extra condition: the 
agent has to interact with the target.

\mypara{State estimation.} We train an object detector using data from the ALFRED simulator that can not only localize and identify all \num{116} ALFRED object classes but also estimate their reachability (\ie, within \num{1.5}m or not). We build upon the Mask R-CNN model~\cite{he2017mask} and introduce an additional binary classification head for the reachability of each detected object. The ground-truth labels for reachability are obtained from the ALFRED simulator for training.

\mypara{Milestone checking.} As mentioned earlier, to reach either a navigation or an interaction milestone, the target objects must be detected and located within a reachable distance. To check this, we compare the target object names, which are extracted from the language instruction (\eg, \nl{kitchen island}), to the class labels (\eg, \fl{countertop}) of the objects detected by Mask R-CNN, essentially a symbol grounding task. We only consider the detected objects that are estimated to be reachable. We apply an off-the-shelf word similarity tool based on Wordnet~\cite{wordnet} with WUP \cite{wu-palmer-1994-verb} similarity from NLTK \cite{Loper02nltk} to match the target names with the object labels. The reachable object whose label has the highest similarity (above a threshold) to a milestone target is considered as the \emph{grounded instance} of that target; the target is then marked as a success.

For interaction milestones, we need to further check if the agent is interacting/has interacted with the target. As defined in \S\ref{s_background}, an interaction action is a tuple of \fl{(action, object mask)}; the object mask is simply a binary map over the input image. To determine if the agent's action is for the milestone target, we calculate the intersection-over-union (IoU) score between the object mask and the milestone target (provided by Mask R-CNN): if the IoU score is over a certain threshold (\num{0.5}), it is considered matched with the target object of the milestone. For an interaction milestone with multiple targets, the agent has to perform multiple interaction actions to interact with all of them. We keep a checklist of all the milestone targets. The milestone is reached after all the targets have been interacted with.

\subsection{Planning with \milestone}
\label{sec:planning_with_milestone}

The discussion of \milestone so far is detached from the agent. The next question is, \emph{how can \milestone affect an agent's actions, \eg, to prevent it from skipping a subtask?} We propose two simple yet effective ways. First, at any time step, we feed the agent with only the instruction of the current subtask determined by \milestone. This explicitly guides the agent to focus on the current subtask. Specifically in ALFRED, we feed the concatenation of $I_H$ and the one sentence in $I_L$ for the current subtask as opposed to the entire $I_L$. We do so starting from the beginning of a task, when the first sentence of $I_L$ is guaranteed to be the first subtask. We then proceed to the next sentence only after the current subtask is marked as completed by \milestone. The use of \milestone frees the agent from solely relying on its internal mechanism like attention and hidden states to decide subtask switching.

Second, we apply the milestone checker \emph{proactively} for interaction milestones --- before the agent executes its predicted action. This can prevent an agent from interacting with a wrong object, as opposed to trying to correct the mistake after it has happened. For example, if the milestone is \fl{(interaction, fork)} but the agent's predicted binary mask for interaction does not overlap with the grounded instance of fork in the image, \milestone will reject the agent's action by asking it to select another \fl{(action, object mask)} tuple (\autoref{fig_milestone_checking}).
This saves the agent from having to generate an action sequence for recovery, for example, to put the incorrectly picked-up object back down.

In our implementation, if the first interaction action is rejected, we move on to the next action in the agent's top $N$ list (\eg, from a softmax classifier). We iterate over the $N$ actions until we find an interaction action whose mask matches with the milestone target or we find a navigation action instead (\eg, when the right object is not in sight). If none of those happens, the agent will take its top ranked navigation action. We set $N$ to be \num{5} in the experiments.

\subsection{Agent Models}
\label{s_Model}

\subsubsection{\BaseVLNBERT Baseline}
\label{sec:vlnbert}

\begin{figure}
    \centering
    \includegraphics[width=.9\linewidth]{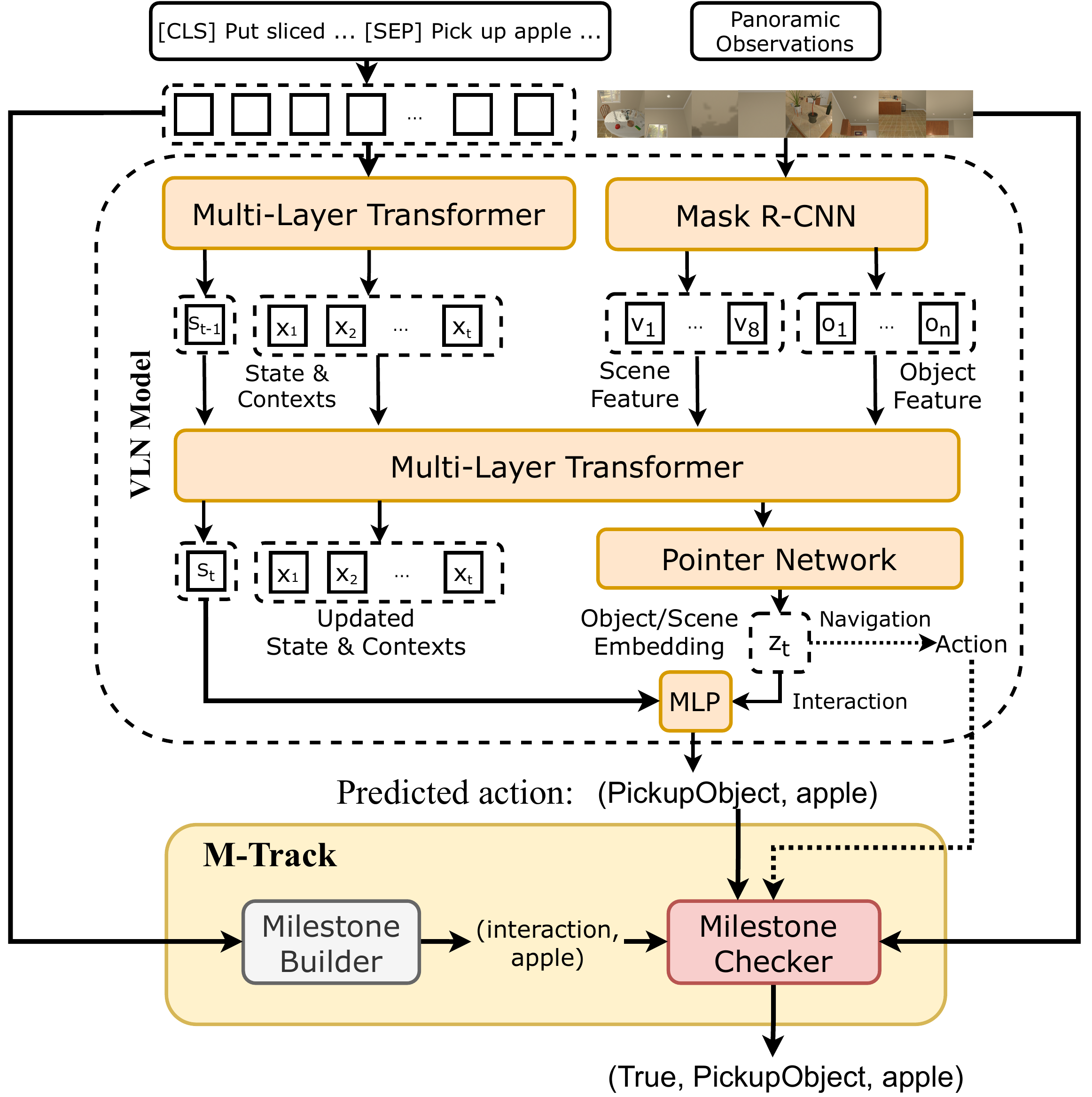}
    \caption{\textbf{Architecture of \BaseVLNBERT with \milestone.}}
    \vskip -10pt
    \label{fig_model}
\end{figure}

Recently, Transformer-based models are becoming increasingly popular for VLN tasks \cite{pashevich2021episodic, pashevich2021episodic, zhang2021hierarchical, suglia2021embodied}. Following this line of work, we build upon the \BaseVLNBERT \cite{Hong_2021_CVPR} model which introduces the concept of recurrent state vector into the Transformer architecture. Since \BaseVLNBERT was designed for the R2R dataset, which contains mostly short-horizon navigation tasks, we adapt it for ALFRED with a series of modifications. Input-wise, we utilize a pre-trained vision encoder\footnote{For simplicity, we use the same Mask R-CNN model that is used in our milestone checker, but it is not necessary.} to extract a scene feature from \num{8} panoramic views and also object features from each view as our visual input. For action prediction, unlike \BaseVLNBERT that only deals with navigation actions, we employ a pointer network \cite{PointerNIPS2015} to choose between navigation, interaction, and stop actions: if the pointer network chooses a scene feature, agent outputs the navigation actions needed to navigate to that scene; if it chooses an object feature, agent outputs the mask for that object, and additionally use an MLP to predict the interaction action type; if it chooses a stop feature (added to the list of visual features as an all-zero vector), agent outputs \fl{Stop}. The MLP takes the concatenation of the chosen object feature and the updated state embedding as input. The architecture as well as its integration with \milestone is illustrated in \autoref{fig_model}, and more implementation details are provided in the supplementary materials.

\subsubsection{\BaseLSTM Baseline}
\label{sec:lstm}

To further show the model-agnostic nature of our \milestone, we use the LSTM baseline introduced in ALFRED \cite{ALFRED20}, and extend the architecture with the same pre-trained vision encoder used in \BaseVLNBERT. Furthermore, to leverage the power of the pre-trained vision encoder, we follow ~\cite{pashevich2021episodic, singh2021factorizing} and ask our agent to select an object from the detected objects instead of directly predicting a binary mask. The corresponding pixel mask is retrieved from the selected object. Refer to supplementary materials for details.

\subsection{Learning}
\label{s_Learning}

As shown in the ALFRED paper~\cite{ALFRED20}, base models like the \BaseLSTM performs rather poorly on ALFRED when simply trained with behavior cloning. Prior studies on other VLN tasks have demonstrated the importance of reinforcement learning (RL) \cite{Hong_2021_CVPR,zhu2020babywalk,tan2019learning}, but its effectiveness has not been validated on ALFRED. We train the models with a combination of behavior cloning (using the cross-entropy loss between the predicted action sequence and the ground truth), object feature selection loss (for interaction actions), and RL.
We apply the A2C algorithm~\cite{pmlr-a2c-mniha16} which, at time $t$, samples an action $a_t$ according to agent's predicted log probability distribution $\log(p^a_t)$, and measures the advantage for that action $adv_t$ with a critic network and a reward. We consider four different types of reward: 1) the straight line distance between the agent and the current navigation/interaction target, 2) the interaction action matching with the ground-truth interaction action which we can compute from the environment state, 3) the visibility of the target, \ie, whether the target is reachable (within 1.5m in ALFRED) and is in sight by the agent, and 4) the final task success. Following VLNBERT~\cite{Hong_2021_CVPR}, we combine behavior cloning loss, cross-entropy loss for object selection and A2C.


\section{Evaluation}
\label{s_exp}

\begin{table*}[t]
    \centering
    \small
    \tabcolsep 10pt
    \renewcommand\arraystretch{0.8}
    \captionsetup{width=.85\textwidth} 
    \begin{tabular}{lcccccc}
    \toprule
    \multirow{2}{*}{\textbf{Model}} &  \multicolumn{3}{c}{\textbf{Test Unseen}} & \multicolumn{3}{c}{\textbf{Test Seen}}\\
    \cmidrule(r){2-4} \cmidrule(r){5-7}
    &  \textbf{SR} & \textbf{PLWSR} & \textbf{GC} & \textbf{SR} & \textbf{PLWSR} & \textbf{GC}\\
    \midrule
    MOCA~\cite{singh2021factorizing}    & 5.30 & 2.72 & 14.28 & 22.05 & 15.10 & 28.29 \\
    LAV~\cite{nottingham2021modular}    & 6.38 & 3.12 & 17.27 & 13.35 & 6.31 & 23.21 \\
    EmBERT~\cite{suglia2021embodied}    & 7.52 & 3.58 & 16.33 & 31.77 & 23.41 & 39.27 \\
    E.T.~\cite{pashevich2021episodic}   & 8.57 & 4.10 & 18.56 & \underline{38.42} & \textbf{27.78} & \underline{45.44} \\
    LWIT~\cite{nguyen2021look}          & 9.42 & 5.60 & 20.91 & 30.92 & \underline{25.90} & 40.53 \\
    HiTUT~\cite{zhang2021hierarchical}  & 13.87 & 5.86 & 20.31 & 21.27 & 11.10 & 29.97 \\
    ABP~\cite{kim2021agent}             & \underline{15.43} & 1.08 & \underline{24.76} & \textbf{44.55} & 3.88 & \textbf{51.13} \\
    HLSM~\cite{blukis2021persistent}    & \textbf{16.29} & 4.34 & \textbf{27.24} & 25.11 & 6.69 & 35.79\\
    \midrule
    \EnhancedLSTM & 8.70 & 4.05 & 16.97 & 14.04 & 7.20 & 21.73 \\
    \LSTMMilestone & 13.28 & \underline{6.25} & 20.20 & 22.05 & 12.83 & 30.48 \\ \hdashline
    \EnhancedVLNBERT & 12.23 & 5.60 & 19.64 & 21.46 & 11.56 & 28.99 \\
    \VLNBERTMilestone & \textbf{16.29} &  \textbf{7.66} & 22.60 & 24.79 & 13.88 &  33.35 \\
    \bottomrule
    \end{tabular}
    \caption{\small \textbf{Performance on the ALFRED test set.} We evaluate \Milestone on \LSTM and \VLNBERT. \Milestone notably improves all evaluation metrics on both test unseen and seen splits. Note that \Milestone using \VLNBERT(or \LSTM) achieves comparable gains to other existing methods. \textbf{Bold} refers to the highest score and \underline{underline} refers to the second highest score.}
    \vskip -10pt
    \label{tab_test}
\end{table*}

\subsection{Experimental Setup}

\mypara{ALFRED.} 
We validate our approach on the ALFRED~\cite{ALFRED20} dataset which evaluates an agent's language-guided navigation and interaction abilities for common household tasks. ALFRED consists of \num{8055} expert demonstrations annotated with \num{25743} natural language instructions. The standard training/validation/test splits contain \num{21023}/\num{1641}/\num{3062} examples, respectively. The validation and test sets are further split into 1) a \textit{seen} set where the environments have been seen during training and 2) an \textit{unseen} set that contains new environments. The validation/test sets include \num{820}/\num{1533} seen and \num{821}/\num{1529} unseen examples.

\mypara{Evaluation metrics.} 
We report the three main metrics used by the ALFRED leaderboard. Success Rate (SR): a binary indicator of whether all subtasks were completed. Path-Length Weighted Success Rate (PLWSR): SR weighted by (expert demonstration path length)/(agent path length). Goal-Condition Success Rate (GC): ratio of completed goal-conditions.\footnote{For example in \cref{fig_alfred}, there are 3 goal conditions: a potato is heated, a potato is on the counterop, and a heated potato is on the counter.} We note that success rate on the unseen test set is considered the primary metric for ranking because models are prone to memorizing the seen environments and often fail to generalize to unseen environments.

\mypara{Models for comparison.}
We denote the base models described in \S\ref{sec:vlnbert} and \S\ref{sec:lstm} as \BaseVLNBERT and \BaseLSTM, respectively. To improve their competence on ALFRED, we further augment them with 1) pre-training their vision encoder on ALFRED images, and 2) reinforcement learning (\S\ref{s_Learning}). We denote the enhanced models as \EnhancedVLNBERT and \EnhancedLSTM, indicating their improved capability for long-horizon tasks. Finally, we integrate each of them with \milestone. 
Even though the focus of our evaluation is to test the effectiveness of \milestone on improving different base models, we still compare our results with other methods that are already published. Please refer to the supplementary materials for implementation details.

\begin{table}[t]
    \centering
    \small
    \tabcolsep 10pt
    \renewcommand{\arraystretch}{0.6}
    \begin{tabular}{ccccc}
    \toprule
    & \multicolumn{2}{c}{\textbf{\LSTM}} & \multicolumn{2}{c}{\textbf{\VLNBERT}} \\
    \cmidrule(r){2-3} \cmidrule(r){4-5}
    \textbf{Train/Test} & $\bm{-}$ & $\bm{+}$ & $\bm{-}$ & $\bm{+}$ \\
    \midrule
    $\bm{-}$ & 9.37 & 10.48 & 10.35 & 13.29 \\
    $\bm{+}$ & 12.20 & \textbf{15.83} & 13.17 & \textbf{17.29} \\
    \bottomrule
    \end{tabular}
    \caption{\small \textbf{Unseen SR on ALFRED validation set with ($\bm{+}$) or without ($\bm{-}$) \milestone during training and/or test.} For example, the \num{17.29} cell indicates when \milestone is integrated into \EnhancedVLNBERT during both training and test.}
    \vskip -10pt
    \label{tab_2x2_unseen_lstm_vln}
\end{table}

\subsection{Main Results}

We summarize the main results on the ALFRED test set in \autoref{tab_test}. First of all, the results show that both of our base models are highly competitive, performing on par or better than many recent VLN models such as E.T., LWIT, and HiTUT. On top of that, \milestone is highly effective in improving both base models: it improves the unseen SR of \EnhancedLSTM and \EnhancedVLNBERT by $\textbf{4.6\%}$ and $\textbf{4.1\%}$ absolute ($\textbf{53\%}$ and $\textbf{33\%}$ relative). Finally, \VLNBERTMilestone performed as much as the best published method, HLSM, on unseen SR (main metric), better on PLWSR for both seen and unseen, similarly on seen SR. The higher seen and unseen PLWSR indicate that our method successfully reduced the path length by focusing on the current subtask to complete the task.

\subsection{Fine-grained Analyses}

\subsubsection{When to Apply \Milestone?}

The flexibility of \milestone makes it possible to be applied at training time, test time, or both. In \autoref{tab_2x2_unseen_lstm_vln}, we show that \milestone is already beneficial when applied only during training or test time, but the gain is most significant when it is applied during both training and test, suggesting that \milestone may be helping the base models in different ways during different phases.

\begin{table*}[t]
    \centering
    \small
    \renewcommand\arraystretch{0.8}
    \captionsetup{width=.95\textwidth}
    \begin{tabular}{lcccccccccc}
\toprule
    \multirow{2}{*}{\textbf{Model}} & \multicolumn{6}{c}{\textbf{Component}} & \multicolumn{2}{c}{\textbf{Val Unseen}} & \multicolumn{2}{c}{\textbf{Val Seen}} \\
    \cmidrule(r){2-7} \cmidrule(r){8-9} \cmidrule(r){10-11}
    & \textbf{RL} & \textbf{ALFRED-OD} & \textbf{Binary} & \textbf{Passive} & \textbf{Proactive} & \textbf{GT} & \textbf{SR} & \textbf{GC} & \textbf{SR} & \textbf{GC} \\
    \midrule
    \textbf{\BaseLSTM}   & & & &  & & &                  1.82    &  3.09 & 9.26  &  11.09  \\ \midrule
     & \cmark & & & & & &                   8.03    & 9.83 & 11.88 & 17.75 \\
     {\footnotesize \EnhancedLSTM} & \cmark & \cmark & &  &  &  &         9.37    & 12.56 & 15.00 & 18.37 \\
     & \cmark & \cmark & \cmark &  & &  &   10.83   & 13.39 & 17.68 & 20.46  \\
      & \cmark & \cmark & & \cmark & &  &    15.22   & 18.88 & 20.97 & 24.73  \\
     {\footnotesize \LSTMMilestone} & \cmark & \cmark & &  & \cmark & &    \textbf{15.83}  & \textbf{20.34} & \textbf{21.70} & \textbf{25.45}   \\
    \hdashline
     \cellcolor{Gray} & \cellcolor{Gray}\cmark & \cellcolor{Gray}\cmark & \cellcolor{Gray} &  \cellcolor{Gray} &  \cellcolor{Gray} & \cellcolor{Gray}\cmark &   \cellcolor{Gray}{20.36}   & \cellcolor{Gray}{30.79} & \cellcolor{Gray}{25.12} & \cellcolor{Gray}{31.41}  \\
    \midrule
    \textbf{\BaseVLNBERT} & & & & & & &                  3.66    & 7.19 & 14.51 & 20.11    \\ \midrule
     & \cmark & & &  & & &                  9.37    & 16.42 & 16.83 & 23.12    \\
     {\footnotesize \EnhancedVLNBERT} & \cmark & \cmark & &  & &  &          10.35   & 18.94 & 21.32 & 25.67    \\
     & \cmark & \cmark & \cmark &  & &  &   14.85   & 22.13 & 22.92 & 28.90    \\
      & \cmark & \cmark & & \cmark & & &     17.05   & 27.37 & 25.48 & 32.07     \\
     {\footnotesize \VLNBERTMilestone} & \cmark & \cmark & &  & \cmark & &    \textbf{17.29} & \textbf{28.98} & \textbf{26.70} & \textbf{33.21}   \\
    \hdashline
     \cellcolor{Gray} & \cellcolor{Gray}\cmark & \cellcolor{Gray}\cmark & \cellcolor{Gray} &  \cellcolor{Gray} &  \cellcolor{Gray} & \cellcolor{Gray}\cmark &   \cellcolor{Gray}{24.38}   & \cellcolor{Gray}{39.34} & \cellcolor{Gray}{31.95} & \cellcolor{Gray}{46.27}     \\
    \bottomrule
    \end{tabular}
    \caption{\small \textbf{Ablation studies on the validation set.} \textbf{RL}: Reinforcement learning. \textbf{ALFRED-OD}: Mask R-CNN object detector pre-trained on ALFRED training images. \textbf{Binary}: Binary milestone classifier. \textbf{Passive}: Milestone checking \textit{after} an action is executed.
    \textbf{Proactive}: Milestone checking \textit{before} an action is executed. \textbf{GT}: \Milestone with ground-truth milestones (an upper bound). 
    }
    \vskip -10pt
    \label{tab_ablation}
\end{table*}

\subsubsection{Ablation Studies}
\label{sec:component_analysis}

\autoref{tab_ablation} shows the effectiveness of different components. 

\mypara{Reinforcement learning.} We show that RL, with our reward design, dramatically improves the performance of both base models, especially in unseen environments. This clearly demonstrates the importance of RL for long-horizon VLN tasks. While similar findings have been discussed for tasks with shorter horizons like R2R~\cite{Hong_2021_CVPR,zhu2020babywalk,tan2019learning}, we are among the first to validate its importance on ALFRED.

\mypara{Pre-training object detector on ALFRED.}
The default Mask R-CNN model is pre-trained on COCO~\cite{lin2014microsoft}. We continue pre-training it on ALFRED (ALFRED-OD), which further improves the performance.

\mypara{Different milestone checking strategies.}
For milestone checking, we compare the passive checking and proactive checking strategies discussed in \S\ref{milestone_checking}. 
As shown in \autoref{tab_ablation}, proactive checking performs better than passive checking, suggesting that \textit{preventing a wrong action from happening is more preferable than correcting the mistake afterwards}. In contrast to our milestone checking, prior work~\cite{das2018embodied} has proposed a binary classifier to check the completeness of current instruction. 
To compare with that, we implement a binary classifier using an MLP conditioned on the hidden state (\LSTM) or the state encoding (\VLNBERT) that predicts if the current milestone has been reached. While it also helps, our milestone checking strategy is still advantageous by a large margin.
Finally, we also estimate an upper bound for \Milestone using ground-truth milestones from the environment instead of our milestone builder. While the results still show a decent room for improvement, the gap is not dramatic, indicating that our milestone builder is reasonably accurate, echoing \autoref{tab_milestone_generation}.



\subsection{Case Studies}
\begin{figure}[t]
    \centering
    \includegraphics[width=1\linewidth]{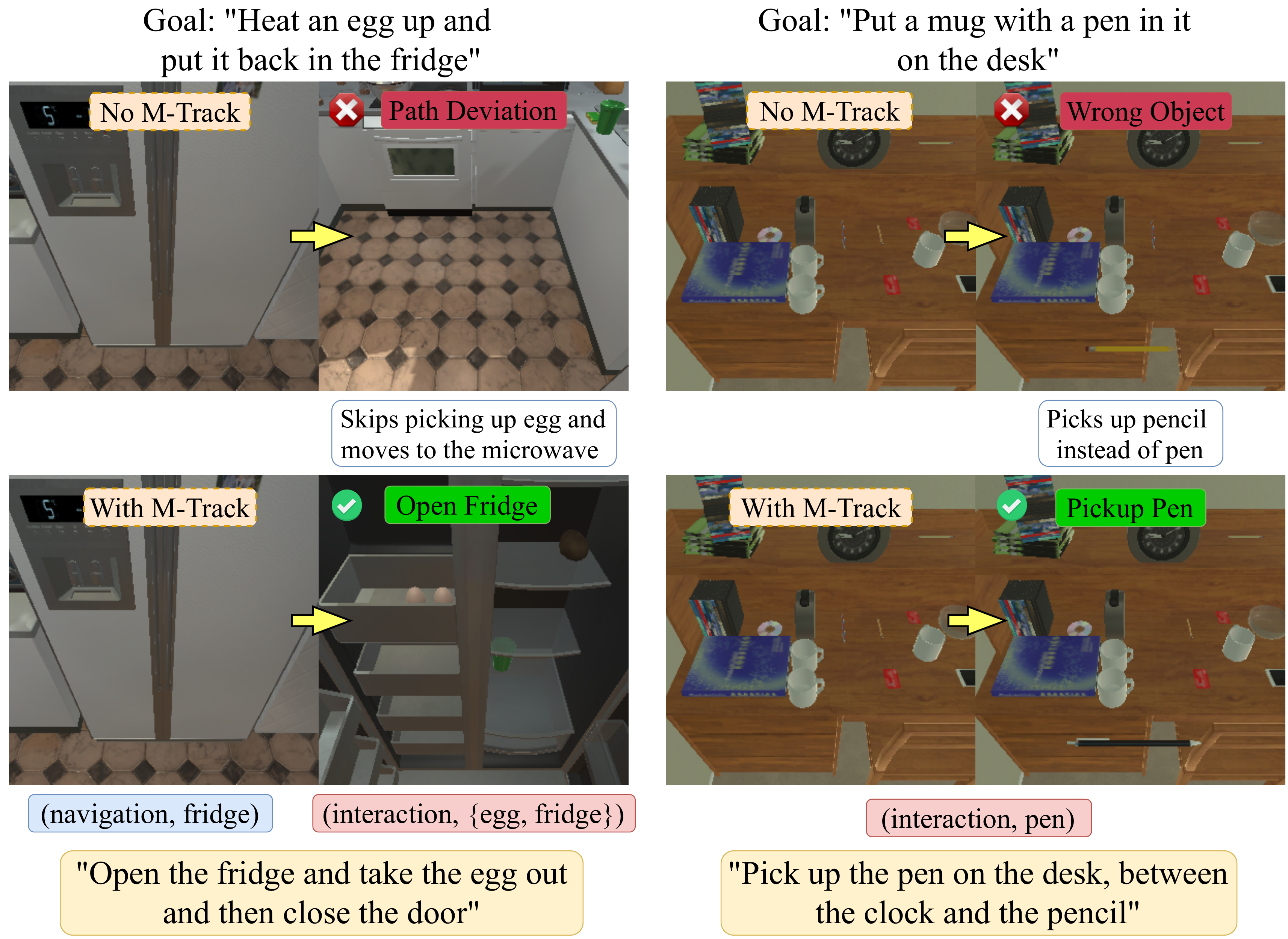}
    \caption{\textbf{Case studies for \milestone}.}
    \label{fig_qual}
    \vskip -10pt
\end{figure}

We compares \EnhancedVLNBERT (top) with \VLNBERTMilestone (bottom) on two validation examples (left and right) to show the importance of \milestone (see \autoref{fig_qual}). 
First, \EnhancedVLNBERT (top-left) skips the current instruction without completing it (\nl{take the egg out from the fridge}), showing its limitation on the long-horizon task. In contrast, \VLNBERTMilestone (bottom-left) completes all subtasks and eventually completes the whole task. Second, \EnhancedVLNBERT (top-right) chooses the wrong object \fl{pencil} instead of the correct object \fl{pen}. The agent may be confused between the two objects since \nl{pencil} also appears in the current instruction and indeed is semantically/visually similar to a pen. In contrast, \VLNBERTMilestone (bottom-right) correctly performs the interaction task because of proactive milestone checking.

\section{Discussion and Conclusions}
\label{s_disc}

We introduce a novel \milestonelong (\milestone) for vision-and-language navigation (VLN) and show that explicit milestone detection and checking significantly benefits long-horizon VLN tasks such as those in ALFRED~\cite{ALFRED20}. Our empirical results show the effectiveness of \milestone with two strong baseline models.
In summary, this work clearly demonstrates the importance of \textit{explicit progress monitoring} (as opposed to, \eg, resorting to a single policy network for both planning and implicit progress monitoring), especially for long-horizon tasks. To make the point, we propose one instantiation with reference to the conditions in ALFRED, and different (or more generic) instantiations for different conditions can be explored in the future. We note the following limitations of the current design that warrant further development:

\noindent \textbf{Assumptions in milestone builder.} Our current instantiation assumes divisible language instructions corresponding to subtasks. It is worth mentioning that prior work (\eg, BabyWalk~\cite{zhu2020babywalk}) does try a similar task decomposition idea on the R2R dataset and shows promising results, and we believe M-Track could be adapted similarly, though it is a less interesting setting for our purpose because of the short horizons. Nonetheless, in more general, realistic settings, accurate milestone building will likely be more challenging, especially when milestones are implicit (\eg, \nl{fetch a cold beer}). One interesting direction is to discover milestones by inductive reasoning over training instances instead of solely from language instructions. Event process mining techniques~\cite{zhang2020analogous}
could potentially be leveraged to discover that \nl{fetch a cold X} generally entails going to a fridge and fetching X (\eg, \nl{beer}) from it.

\noindent \textbf{Assumptions in milestone checker.} 
To date, most VLN tasks are \textit{declarative}. Milestone/goal checking thus can be done by checking solely against the environment state. For \textit{procedural} instructions (\eg, \nl{turn around twice}) the milestone checker may need to check against the agent's action history, though such instructions are rare in existing datasets.

\noindent \textbf{Non-unique golden trajectories.} Though uncommon in ALFRED, in more complex tasks and/or environments, there could exist multiple viable trajectories (\eg, different execution orders of subtasks leading to the same goal state) to complete a task. Currently milestones are assumed to be hard constraints that an agent has to achieve in order to proceed. It may be helpful to (learn to) soften the constraints imposed by milestones to provide more flexibility.

\section*{Acknowledgement}

The authors would like to thank the colleagues from the OSU NLP group for their thoughtful comments. This research was supported in part by NSF OAC 2118240 and NSF OAC 2112606. J.\ Kil, T.\ Pan, and W.\ Chao are also partially supported by NSF IIS 2107077 and the OSU GI Development funds.

{\small
\bibliographystyle{ieee_fullname}
\bibliography{egbib}

\begin{thebibliography}{10}\itemsep=-1pt

\bibitem{anderson2018vision}
Peter Anderson, Qi Wu, Damien Teney, Jake Bruce, Mark Johnson, Niko
  S{\"u}nderhauf, Ian Reid, Stephen Gould, and Anton Van Den~Hengel.
\newblock Vision-and-language navigation: Interpreting visually-grounded
  navigation instructions in real environments.
\newblock In {\em CVPR}, 2018.

\bibitem{blukis2018following}
Valts Blukis, Nataly Brukhim, Andrew Bennett, Ross~A Knepper, and Yoav Artzi.
\newblock Following high-level navigation instructions on a simulated
  quadcopter with imitation learning.
\newblock {\em arXiv preprint arXiv:1806.00047}, 2018.

\bibitem{blukis2020few}
Valts Blukis, Ross~A Knepper, and Yoav Artzi.
\newblock Few-shot object grounding and mapping for natural language robot
  instruction following.
\newblock {\em arXiv preprint arXiv:2011.07384}, 2020.

\bibitem{blukis2021persistent}
Valts Blukis, Chris Paxton, Dieter Fox, Animesh Garg, and Yoav Artzi.
\newblock A persistent spatial semantic representation for high-level natural
  language instruction execution.
\newblock {\em arXiv preprint arXiv:2107.05612}, 2021.

\bibitem{chen2019touchdown}
Howard Chen, Alane Suhr, Dipendra Misra, Noah Snavely, and Yoav Artzi.
\newblock Touchdown: Natural language navigation and spatial reasoning in
  visual street environments.
\newblock In {\em CVPR}, 2019.

\bibitem{chen2019behavioral}
Kevin Chen, Juan~Pablo de Vicente, Gabriel Sepulveda, Fei Xia, Alvaro Soto,
  Marynel V{\'a}zquez, and Silvio Savarese.
\newblock A behavioral approach to visual navigation with graph localization
  networks.
\newblock {\em arXiv preprint arXiv:1903.00445}, 2019.

\bibitem{das2018embodied}
Abhishek Das, Samyak Datta, Georgia Gkioxari, Stefan Lee, Devi Parikh, and
  Dhruv Batra.
\newblock Embodied question answering.
\newblock In {\em CVPR}, 2018.

\bibitem{das2018neural}
Abhishek Das, Georgia Gkioxari, Stefan Lee, Devi Parikh, and Dhruv Batra.
\newblock Neural modular control for embodied question answering.
\newblock In {\em Conference on Robot Learning}, 2018.

\bibitem{devlin2018bert}
Jacob Devlin, Ming-Wei Chang, Kenton Lee, and Kristina Toutanova.
\newblock Bert: Pre-training of deep bidirectional transformers for language
  understanding.
\newblock {\em arXiv preprint arXiv:1810.04805}, 2018.

\bibitem{wordnet}
Christiane Fellbaum.
\newblock {\em WordNet: An Electronic Lexical Database}.
\newblock Bradford Books, 1998.

\bibitem{fried2018speaker}
Daniel Fried, Ronghang Hu, Volkan Cirik, Anna Rohrbach, Jacob Andreas,
  Louis-Philippe Morency, Taylor Berg-Kirkpatrick, Kate Saenko, Dan Klein, and
  Trevor Darrell.
\newblock Speaker-follower models for vision-and-language navigation.
\newblock {\em arXiv preprint arXiv:1806.02724}, 2018.

\bibitem{gordon2018iqa}
Daniel Gordon, Aniruddha Kembhavi, Mohammad Rastegari, Joseph Redmon, Dieter
  Fox, and Ali Farhadi.
\newblock Iqa: Visual question answering in interactive environments.
\newblock In {\em CVPR}, 2018.

\bibitem{hao2020towards}
Weituo Hao, Chunyuan Li, Xiujun Li, Lawrence Carin, and Jianfeng Gao.
\newblock Towards learning a generic agent for vision-and-language navigation
  via pre-training.
\newblock In {\em CVPR}, 2020.

\bibitem{he2017mask}
Kaiming He, Georgia Gkioxari, Piotr Doll{\'a}r, and Ross Girshick.
\newblock Mask r-cnn.
\newblock In {\em ICCV}, 2017.

\bibitem{he2016deep}
Kaiming He, Xiangyu Zhang, Shaoqing Ren, and Jian Sun.
\newblock Deep residual learning for image recognition.
\newblock In {\em CVPR}, 2016.

\bibitem{hong2020language}
Yicong Hong, Cristian Rodriguez-Opazo, Yuankai Qi, Qi Wu, and Stephen Gould.
\newblock Language and visual entity relationship graph for agent navigation.
\newblock {\em arXiv preprint arXiv:2010.09304}, 2020.

\bibitem{hong2020sub}
Yicong Hong, Cristian Rodriguez-Opazo, Qi Wu, and Stephen Gould.
\newblock Sub-instruction aware vision-and-language navigation.
\newblock {\em arXiv preprint arXiv:2004.02707}, 2020.

\bibitem{Hong_2021_CVPR}
Yicong Hong, Qi Wu, Yuankai Qi, Cristian Rodriguez-Opazo, and Stephen Gould.
\newblock A recurrent vision-and-language bert for navigation.
\newblock In {\em CVPR}, 2021.

\bibitem{jain2019stay}
Vihan Jain, Gabriel Magalhaes, Alexander Ku, Ashish Vaswani, Eugene Ie, and
  Jason Baldridge.
\newblock Stay on the path: Instruction fidelity in vision-and-language
  navigation.
\newblock {\em arXiv preprint arXiv:1905.12255}, 2019.

\bibitem{ke2019tactical}
Liyiming Ke, Xiujun Li, Yonatan Bisk, Ari Holtzman, Zhe Gan, JJ~(Jingjing) Liu,
  Jianfeng Gao, Yejin Choi, and Siddhartha Srinivasa.
\newblock Tactical rewind: Self-correction via backtracking in
  vision-and-language navigation.
\newblock In {\em CVPR}, 2019.

\bibitem{kim2021agent}
Byeonghwi Kim, Suvaansh Bhambri, Kunal~Pratap Singh, Roozbeh Mottaghi, and
  Jonghyun Choi.
\newblock Agent with the big picture: Perceiving surroundings for interactive
  instruction following.
\newblock In {\em Embodied AI Workshop CVPR}, 2021.

\bibitem{ai2thor}
Eric Kolve, Roozbeh Mottaghi, Winson Han, Eli VanderBilt, Luca Weihs, Alvaro
  Herrasti, Daniel Gordon, Yuke Zhu, Abhinav Gupta, and Ali Farhadi.
\newblock {AI2-THOR: An Interactive 3D Environment for Visual AI}.
\newblock {\em arXiv}, 2017.

\bibitem{kurita2020generative}
Shuhei Kurita and Kyunghyun Cho.
\newblock Generative language-grounded policy in vision-and-language navigation
  with bayes' rule.
\newblock {\em arXiv preprint arXiv:2009.07783}, 2020.

\bibitem{crf}
John~D. Lafferty, Andrew McCallum, and Fernando C.~N. Pereira.
\newblock Conditional random fields: Probabilistic models for segmenting and
  labeling sequence data.
\newblock In {\em Proceedings of the Eighteenth International Conference on
  Machine Learning}, ICML '01, page 282–289, San Francisco, CA, USA, 2001.
  Morgan Kaufmann Publishers Inc.

\bibitem{li2021improving}
Jialu Li, Hao Tan, and Mohit Bansal.
\newblock Improving cross-modal alignment in vision language navigation via
  syntactic information.
\newblock {\em arXiv preprint arXiv:2104.09580}, 2021.

\bibitem{li2020unsupervised}
Juncheng Li, Xin Wang, Siliang Tang, Haizhou Shi, Fei Wu, Yueting Zhuang, and
  William~Yang Wang.
\newblock Unsupervised reinforcement learning of transferable meta-skills for
  embodied navigation.
\newblock In {\em CVPR}, 2020.

\bibitem{li2019robust}
Xiujun Li, Chunyuan Li, Qiaolin Xia, Yonatan Bisk, Asli Celikyilmaz, Jianfeng
  Gao, Noah Smith, and Yejin Choi.
\newblock Robust navigation with language pretraining and stochastic sampling.
\newblock {\em arXiv preprint arXiv:1909.02244}, 2019.

\bibitem{lin2014microsoft}
Tsung-Yi Lin, Michael Maire, Serge Belongie, James Hays, Pietro Perona, Deva
  Ramanan, Piotr Doll{\'a}r, and C~Lawrence Zitnick.
\newblock Microsoft coco: Common objects in context.
\newblock In {\em ECCV}, 2014.

\bibitem{Loper02nltk}
Edward Loper and Steven Bird.
\newblock Nltk: The natural language toolkit.
\newblock In {\em Proceedings of the ACL Workshop on Effective Tools and
  Methodologies for Teaching Natural Language Processing and Computational
  Linguistics. Philadelphia: Association for Computational Linguistics}, 2002.

\bibitem{ma2019self}
Chih-Yao Ma, Jiasen Lu, Zuxuan Wu, Ghassan AlRegib, Zsolt Kira, Richard Socher,
  and Caiming Xiong.
\newblock Self-monitoring navigation agent via auxiliary progress estimation.
\newblock {\em arXiv preprint arXiv:1901.03035}, 2019.

\bibitem{Ma_2019_CVPR}
Chih-Yao Ma, Zuxuan Wu, Ghassan AlRegib, Caiming Xiong, and Zsolt Kira.
\newblock The regretful agent: Heuristic-aided navigation through progress
  estimation.
\newblock In {\em CVPR}, 2019.

\bibitem{majumdar2020improving}
Arjun Majumdar, Ayush Shrivastava, Stefan Lee, Peter Anderson, Devi Parikh, and
  Dhruv Batra.
\newblock Improving vision-and-language navigation with image-text pairs from
  the web.
\newblock In {\em ECCV}, 2020.

\bibitem{mayo2021visual}
Bar Mayo, Tamir Hazan, and Ayellet Tal.
\newblock Visual navigation with spatial attention.
\newblock In {\em CVPR}, 2021.

\bibitem{PDDL}
Drew McDermott, Malik Ghallab, Adele~E. Howe, Craig~A. Knoblock, Ashwin Ram,
  Manuela~M. Veloso, Daniel~S. Weld, and David~E. Wilkins.
\newblock Pddl-the planning domain definition language.
\newblock 1998.

\bibitem{min2021film}
So~Yeon Min, Devendra~Singh Chaplot, Pradeep Ravikumar, Yonatan Bisk, and
  Ruslan Salakhutdinov.
\newblock Film: Following instructions in language with modular methods.
\newblock {\em arXiv preprint arXiv:2110.07342}, 2021.

\bibitem{misra2018mapping}
Dipendra Misra, Andrew Bennett, Valts Blukis, Eyvind Niklasson, Max Shatkhin,
  and Yoav Artzi.
\newblock Mapping instructions to actions in 3d environments with visual goal
  prediction.
\newblock {\em arXiv preprint arXiv:1809.00786}, 2018.

\bibitem{pmlr-a2c-mniha16}
Volodymyr Mnih, Adria~Puigdomenech Badia, Mehdi Mirza, Alex Graves, Timothy
  Lillicrap, Tim Harley, David Silver, and Koray Kavukcuoglu.
\newblock Asynchronous methods for deep reinforcement learning.
\newblock In {\em ICML}, 2016.

\bibitem{nguyen2021look}
Van-Quang Nguyen, Masanori Suganuma, and Takayuki Okatani.
\newblock Look wide and interpret twice: Improving performance on interactive
  instruction-following tasks.
\newblock {\em arXiv preprint arXiv:2106.00596}, 2021.

\bibitem{nottingham2021modular}
Kolby Nottingham, Litian Liang, Daeyun Shin, Charless~C Fowlkes, Roy Fox, and
  Sameer Singh.
\newblock Modular framework for visuomotor language grounding.
\newblock {\em arXiv preprint arXiv:2109.02161}, 2021.

\bibitem{pashevich2021episodic}
Alexander Pashevich, Cordelia Schmid, and Chen Sun.
\newblock Episodic transformer for vision-and-language navigation.
\newblock {\em arXiv preprint arXiv:2105.06453}, 2021.

\bibitem{puig2018virtualhome}
Xavier Puig, Kevin Ra, Marko Boben, Jiaman Li, Tingwu Wang, Sanja Fidler, and
  Antonio Torralba.
\newblock Virtualhome: Simulating household activities via programs.
\newblock In {\em CVPR}, 2018.

\bibitem{regneri2013grounding}
Michaela Regneri, Marcus Rohrbach, Dominikus Wetzel, Stefan Thater, Bernt
  Schiele, and Manfred Pinkal.
\newblock Grounding action descriptions in videos.
\newblock {\em In TACL}, 2013.

\bibitem{ren2015faster}
Shaoqing Ren, Kaiming He, Ross Girshick, and Jian Sun.
\newblock Faster r-cnn: Towards real-time object detection with region proposal
  networks.
\newblock {\em NeurIPS}, 2015.

\bibitem{ILSVRC15}
Olga Russakovsky, Jia Deng, Hao Su, Jonathan Krause, Sanjeev Satheesh, Sean Ma,
  Zhiheng Huang, Andrej Karpathy, Aditya Khosla, Michael Bernstein,
  Alexander~C. Berg, and Li Fei-Fei.
\newblock {ImageNet Large Scale Visual Recognition Challenge}.
\newblock {\em IJCV}, 2015.

\bibitem{savinov2018semi}
Nikolay Savinov, Alexey Dosovitskiy, and Vladlen Koltun.
\newblock Semi-parametric topological memory for navigation.
\newblock {\em arXiv preprint arXiv:1803.00653}, 2018.

\bibitem{ALFRED20}
Mohit Shridhar, Jesse Thomason, Daniel Gordon, Yonatan Bisk, Winson Han,
  Roozbeh Mottaghi, Luke Zettlemoyer, and Dieter Fox.
\newblock {ALFRED: A Benchmark for Interpreting Grounded Instructions for
  Everyday Tasks}.
\newblock In {\em CVPR}, 2020.

\bibitem{singh2021factorizing}
Kunal~Pratap Singh, Suvaansh Bhambri, Byeonghwi Kim, Roozbeh Mottaghi, and
  Jonghyun Choi.
\newblock Factorizing perception and policy for interactive instruction
  following.
\newblock In {\em CVPR}, 2021.

\bibitem{Souza2019PortugueseNE}
F{\'a}bio Souza, Rodrigo Nogueira, and Roberto de Alencar~Lotufo.
\newblock Portuguese named entity recognition using bert-crf.
\newblock {\em ArXiv}, abs/1909.10649, 2019.

\bibitem{suglia2021embodied}
Alessandro Suglia, Qiaozi Gao, Jesse Thomason, Govind Thattai, and Gaurav
  Sukhatme.
\newblock Embodied bert: A transformer model for embodied, language-guided
  visual task completion.
\newblock {\em arXiv preprint arXiv:2108.04927}, 2021.

\bibitem{tan2019learning}
Hao Tan, Licheng Yu, and Mohit Bansal.
\newblock Learning to navigate unseen environments: Back translation with
  environmental dropout.
\newblock {\em arXiv preprint arXiv:1904.04195}, 2019.

\bibitem{vaswani2017attention}
Ashish Vaswani, Noam Shazeer, Niki Parmar, Jakob Uszkoreit, Llion Jones,
  Aidan~N Gomez, {\L}ukasz Kaiser, and Illia Polosukhin.
\newblock Attention is all you need.
\newblock In {\em NeurIPS}, 2017.

\bibitem{PointerNIPS2015}
Oriol Vinyals, Meire Fortunato, and Navdeep Jaitly.
\newblock Pointer networks.
\newblock In {\em NeurIPS}, 2015.

\bibitem{wang2021structured}
Hanqing Wang, Wenguan Wang, Wei Liang, Caiming Xiong, and Jianbing Shen.
\newblock Structured scene memory for vision-language navigation.
\newblock In {\em CVPR}, 2021.

\bibitem{wang2020active}
Hanqing Wang, Wenguan Wang, Tianmin Shu, Wei Liang, and Jianbing Shen.
\newblock Active visual information gathering for vision-language navigation.
\newblock In {\em ECCV}, 2020.

\bibitem{wijmans2019embodied}
Erik Wijmans, Samyak Datta, Oleksandr Maksymets, Abhishek Das, Georgia
  Gkioxari, Stefan Lee, Irfan Essa, Devi Parikh, and Dhruv Batra.
\newblock Embodied question answering in photorealistic environments with point
  cloud perception.
\newblock In {\em CVPR}, 2019.

\bibitem{wu-palmer-1994-verb}
Zhibiao Wu and Martha Palmer.
\newblock Verb semantics and lexical selection.
\newblock In {\em ACL}, 1994.

\bibitem{xiang-etal-2020-learning}
Jiannan Xiang, Xin Wang, and William~Yang Wang.
\newblock Learning to stop: A simple yet effective approach to urban
  vision-language navigation.
\newblock In {\em Findings of the Association for Computational Linguistics:
  EMNLP 2020}, pages 699--707, Online, Nov. 2020. Association for Computational
  Linguistics.

\bibitem{xiang2020learning}
Jiannan Xiang, Xin~Eric Wang, and William~Yang Wang.
\newblock Learning to stop: A simple yet effective approach to urban
  vision-language navigation.
\newblock {\em arXiv preprint arXiv:2009.13112}, 2020.

\bibitem{yang2018visual}
Wei Yang, Xiaolong Wang, Ali Farhadi, Abhinav Gupta, and Roozbeh Mottaghi.
\newblock Visual semantic navigation using scene priors.
\newblock {\em arXiv preprint arXiv:1810.06543}, 2018.

\bibitem{zhang2020analogous}
Hongming Zhang and et al.
\newblock Analogous process structure induction for sub-event sequence
  prediction.
\newblock In {\em EMNLP}, 2020.

\bibitem{zhang2021hierarchical}
Yichi Zhang and Joyce Chai.
\newblock Hierarchical task learning from language instructions with unified
  transformers and self-monitoring.
\newblock {\em arXiv preprint arXiv:2106.03427}, 2021.

\bibitem{zhu2020vision}
Fengda Zhu, Yi Zhu, Xiaojun Chang, and Xiaodan Liang.
\newblock Vision-language navigation with self-supervised auxiliary reasoning
  tasks.
\newblock In {\em CVPR}, 2020.

\bibitem{zhu2020babywalk}
Wang Zhu, Hexiang Hu, Jiacheng Chen, Zhiwei Deng, Vihan Jain, Eugene Ie, and
  Fei Sha.
\newblock Babywalk: Going farther in vision-and-language navigation by taking
  baby steps.
\newblock {\em arXiv preprint arXiv:2005.04625}, 2020.

\bibitem{zhu2017visual}
Yuke Zhu, Daniel Gordon, Eric Kolve, Dieter Fox, Li Fei-Fei, Abhinav Gupta,
  Roozbeh Mottaghi, and Ali Farhadi.
\newblock Visual semantic planning using deep successor representations.
\newblock In {\em ICCV}, 2017.

\end{thebibliography}
}

\clearpage

\appendix
\section*{Appendices}
In this supplementary material, we provide details omitted in the main text.
\begin{itemize}
    \item \autoref{s:related}: More related work
    \item \autoref{s:milestone}: \Milestone implementation details
    \item \autoref{s:model}: Model implementation details
    \item \autoref{s:exp}: Additional experiments
\end{itemize}

\section{More Related Work}
\label{s:related}
Due to the space constraint, we only include the most related works in the main text. Here, we add some extra related works to show recent trends in VLN.

\mypara{Auxiliary information in VLN.} \Milestone can be viewed as an auxiliary information to the agent. A variety of auxiliary information has indeed been explored to improve the VLN models~\cite{mayo2021visual,yang2018visual,blukis2021persistent,wang2021structured,savinov2018semi,chen2019behavioral,zhu2020vision,li2021improving,li2020unsupervised,blukis2020few,blukis2018following,xiang2020learning,min2021film}. Several prior works proposed to build a semantic map that encodes the spatial semantic information to bridge the gap between instructions and visual observations~\cite{mayo2021visual,blukis2021persistent,blukis2018following,min2021film}. Other studies suggested a topological map that memorizes previous actions and locations to facilitate planning~\cite{wang2021structured,savinov2018semi,chen2019behavioral}. \milestone is different from them by its simplicity, functionality, and compatibility --- it is completely detached from VLN models and thus model-agnostic.

\mypara{Data Augmentation in VLN.} A number of prior studies investigate data augmentation to increase generalizability in unseen environments. One stream of works focuses on generating synthetic language instructions~\cite{fried2018speaker,pashevich2021episodic,wang2020active,hong2020sub,majumdar2020improving}. E.T.~\cite{pashevich2021episodic} constructs synthetic instructions using the expert path planner in ALFRED. Speaker-Follower~\cite{fried2018speaker} generates human-like textual instructions based on a VLN model trained on ground-truth routes. The other stream considers augmenting visual observations. Most of these works include surrounding views to enlarge an agent's field of view and thus enhance its navigation ability~\cite{kim2021agent,hong2020language,nguyen2021look,blukis2021persistent,suglia2021embodied,hao2020towards}. In line with these studies, we augment visual observations using panoramic views with different angles and headings.

\mypara{Learning Strategies in VLN.}
Several studies train VLN models with imitation learning~\cite{ALFRED20,pashevich2021episodic,anderson2018vision,kurita2020generative,xiang2020learning,li2019robust,singh2021factorizing} while some other works apply reinforcement learning~\cite{li2020unsupervised,yang2018visual,misra2018mapping}. To balance exploration and exploitation in navigation, some recent works leverage both imitation learning and reinforcement learning~\cite{Hong_2021_CVPR,zhu2020babywalk,tan2019learning,li2021improving,jain2019stay,hong2020language}. Following the recent studies, \Milestone exploits both of them and shows notable improvement on ALFRED.

\mypara{Visual Input.} There has been significant recent progress in learning visual representations of views. Several studies take image features encoded by ResNet~\cite{he2016deep} as visual input~\cite{ALFRED20,anderson2018vision,fried2018speaker,zhu2020babywalk}. \BaseVLNBERT~\cite{majumdar2020improving} take as input object features from Faster R-CNN~\cite{ren2015faster} to encode objects' semantic information. Some other studies leverage both image and object features to learn better visual representations~\cite{Hong_2021_CVPR,hong2020language}. Recently, several papers use a pre-trained segmentation model (\eg, Mask R-CNN~\cite{he2017mask}) to obtain more accurate object information~\cite{min2021film,singh2021factorizing,kim2021agent,zhang2021hierarchical,pashevich2021episodic,suglia2021embodied}. Following the recent trend in VLN, \Milestone exploits Mask R-CNN to detect objects for milestone checking and encode their visual representations to facilitate interaction tasks.

\section{\Milestone Implementation Details}
\label{s:milestone}

\subsection{Milestone Builder}

To estimate an upper bound of \milestone, we first build a ground-truth dataset using ground-truth tags derived from the ALFRED~\cite{ALFRED20} expert demonstrations.
The ALFRED expert demonstrations are encoded in Planning Domain Definition Language (PDDL)~\cite{PDDL} rules. PDDL annotations include task-specific goal conditions for each low-level instruction. Each low-level instruction in PDDL language is defined by \fl{(d, i, p)}, where \fl{d = (action, argument)} is a discrete action tuple containing the description of the action and its argument (object), \fl{i} is the index of the low-level instruction, \fl{p = (action, location/ObjectID)} is a planner action which is an action tuple directly applicable to the simulator. We use the discrete action tuple to tag the low-level instruction with the ground-truth object labels. For example, \nl{Go to the trash can on the far side of the kitchen}, is labeled with the discrete action \fl{(GotoLocation, trashcan)}, based on which we automatically tag the instruction as \fl{(O, O, O, B-Nav, I-Nav, O, O, O, O, O, O, O)}. We apply BIO tagging format\footnote{B- prefix indicates the tag is the beginning of an object label, I- prefix indicates the tag is inside/end of the object label, and O refers to all non-tagged words.} to turn the object labels into tags. Every ALFRED low-level instruction has annotated labels, enabling us to build the ground-truth milestone training data easily. After tagging the ALFRED training and validation data with the ground truth object labels in this way, we train a BERT-CRF~\cite{devlin2018bert, Souza2019PortugueseNE} tagger on the training data to predict the milestone tags in the instruction. We choose BERT to utilize its powerful context encoding capability and add a CRF~\cite{crf} layer on top of BERT to better model the interdependence of tag predictions.
BERT-CRF is trained end to end with training data generated from the training split of ALFRED and validated with validation set generated with validation seen and unseen annotations for ALFRED. The model that has the highest F1 on the validation set is chosen.

During the main model (\eg, \VLNBERTMilestone) execution, BERT-CRF outputs tags for a given low-level instruction. For instance, in the instruction, \nl{Turn and go to the sink}, BERT-CRF outputs \fl{(O, O, O, O, O, B-Nav)}. The \fl{(tag, word)} pair is our predicted milestone for the instruction. 
We freeze BERT-CRF during the main model training.

\begin{figure}
    \centering
    \includegraphics[width=1\linewidth]{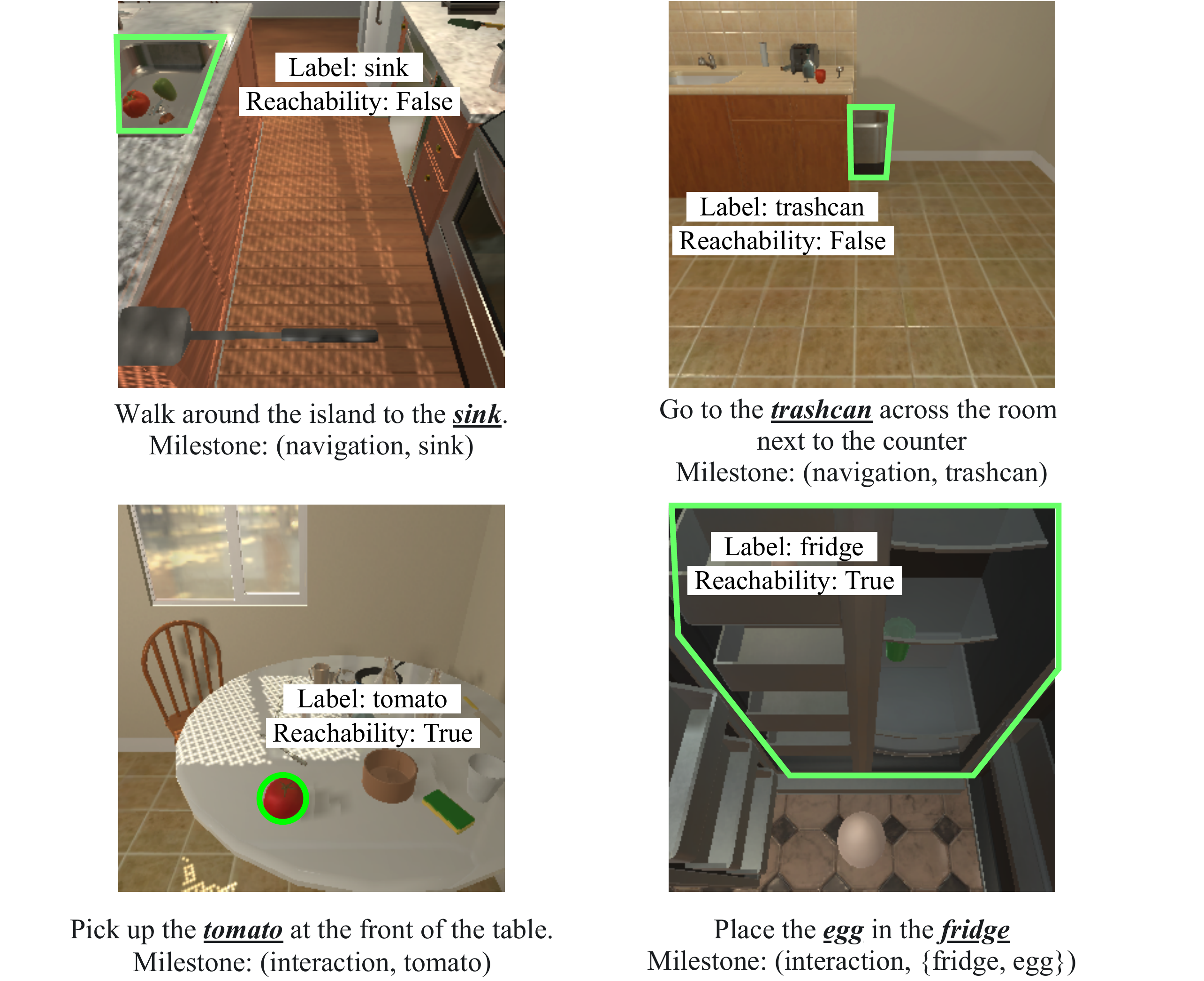}
    \vskip -5pt
    \caption{\textbf{Examples of milestones.}}
    \label{fig:milestones}
\end{figure}

\subsection{Milestone Checking}
Here we elaborate on the design decisions for some corner cases during milestone checking.
\begin{itemize}
    \item \textbf{What if an instruction contains multiple milestones?}
    \begin{itemize}
        \item[1.] Navigation + Navigation: Navigate to the first milestone then to the next.
        \item[2.] Navigation + Interaction: Navigate to the navigation milestone first and navigate/interact with the interaction milestone.
        \item[3.] Interaction + Interaction: Interact with both objects without specifying a fixed order.
    \end{itemize}
    \item \textbf{What if an instruction contains no milestone?} 
    There are only less than 1\% of such cases in the validation unseen environments in ALFRED (\ie, \num{40} out of \num{5140} instructions). For those cases, we concatenate the current instruction with the next instruction that has milestones detected.
    \item \textbf{What happens if the milestone builder makes a mistake, \eg, missing a milestone / extracting an unnecessary milestone / detecting a wrong object?}
    Generally the milestone builder is pretty accurate (\textit{c.f.}\ Table 1 in the main paper), so mistakes are not common. However, in the cases when a mistake does occur, we currently have the agent skip the milestone if the checking fails \num{15} consecutive times. On the other hand, if the object detector fails to detect the object, usually the failure is recovered later when the agent is in a different pose to take a new view of the object.
\end{itemize}

\section{Model Implementation Details} 
\label{s:model}
\subsection{\BaseVLNBERT}

Our modified \BaseVLNBERT~\cite{Hong_2021_CVPR} consists of four modules: a language encoder, a vision encoder, an action decoder, and a pointer network with a multi-layer perceptron. 
Given a time step $t$, the language encoder takes current instruction (\ie, the concatenation of the high-level instruction and the current low-level instruction) as input and outputs the contextualized token embeddings. Following \BaseVLNBERT, we consider the \fl{[CLS]} embedding as a state embedding $s_{t}$ representing an agent's current state and denotes other textual token embedding as $x_{i}$\footnote{We omit $t$ on other variables (\eg, textual token embeddings, scene/object features, etc.) for simplicity except for the state embedding.}.
For the vision encoder, we leverage Mask R-CNN to obtain two types of visual features: 1) a scene feature $v_{j}$ representing a view, and 2) an object feature $o_{k}$ indicating an object in the view. In total, we extract 8 scene features from panoramic views (4 headings of 90$^{\circ}$ and 2 elevation angles of $\pm30^{\circ}$) and 20 highest scoring object features from all scenes.
The action decoder then performs a grounded language learning by taking four inputs: a previous state embedding $s_{t-1}$, a sequence of textual embeddings $\{x_{i}\}$, a sequence of scene features $\{v_{j}\}$, and a sequence of object features $\{o_{k}\}$. 
\begin{equation}
\begin{aligned}
   s_t, {} &  \{x'_{i}\}, \{v'_{j}\}, \{o'_{k}\} = \\ & \text{\BaseVLNBERT}(s_{t-1}, \{x_{i}\}, \{v_{j}\}, \{o_{k}\})
\end{aligned}
\end{equation}
Unlike \BaseVLNBERT, we employ a pointer network \cite{PointerNIPS2015} to let the agent choose between navigation and interaction action. The pointer network predicts an action for the time step $t$ by \eqref{pt_un}, \eqref{pt_n_hat}, and \eqref{pt_a}.
\begin{equation}
\label{pt_un}
\begin{aligned}
    u_{n} {} & = z^\top \text{tanh}(W_1 e_{n} + W_2 s_t), \\ &
    e_{n} \in \{v'_{j}\} \cup \{o'_{k}\},  n \in (1,\cdots,J+K)
\end{aligned}
\end{equation}
where $z$, $W_1$, $W_2$ are learnable parameters, $e_{n}$ is either the scene feature or the object feature, and $s_t$ is the updated state embedding.
\begin{equation}
\label{pt_n_hat}
    \hat{n}=\argmax_{n\in(1, \cdots, J+K)}\sigma(u)_{n}
\end{equation}
where $\sigma$ is the softmax normalizing the vector $u$. If $\hat{n}$ represents a scene, the agent should navigate to the scene $\hat{n}$ at the time step $t$. In contrast, if $\hat{n}$ indicates an object, the agent should interact with the object $\hat{n}$ at the time step $t$ by the corresponding interaction action $\hat{a}$
\begin{equation}
\label{pt_a}
    \hat{a} = \argmax_{a \in \text{IA}} \sigma(W_3[s_t;o'_{\hat{n}}])_{a}
\end{equation}
where $W_3$ is the learnable parameter, $o'_{\hat{n}}$ is the feature of object $\hat{n}$, and \text{IA} is the set of 7 interaction actions.

\subsection{\BaseLSTM}
\begin{figure}
    \centering
    \includegraphics[width=1\linewidth]{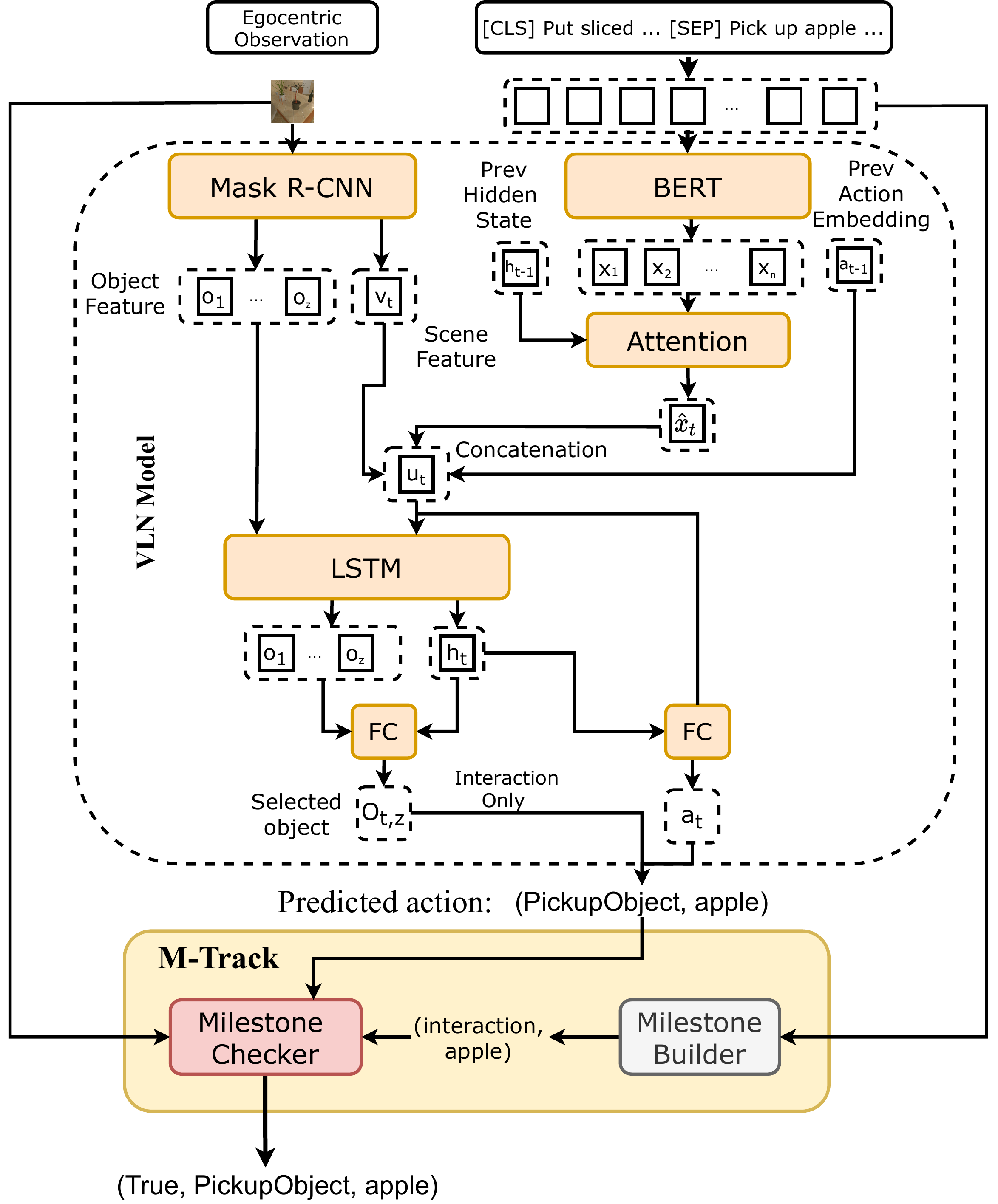}
    \vskip -5pt
    \caption{\textbf{Architecture of \EnhancedLSTM with \Milestone.}}
    \label{fig:LSTM}
\end{figure}

We modify the CNN-LSTM model presented in ALFRED~\cite{ALFRED20} with support for a pre-trained object encoder inspired by common VLN models' approach~\cite{fried2018speaker, tan2019learning, wang2021structured} in choosing between multiple scene features. Given a high level goal instruction $G$ and step-by-step instructions $S = \{s_1, s_2, \cdots s_n\}$ of $n$ instruction sentences, we concatenate the goal instruction with only the relevant step-by-step instruction such as $L = \{G, \text{$<$SEP$>$}, s_n\}$ with $<$SEP$>$ token indicating the difference between the goal and step-by-step instruction. Then we perform a soft-attention on the language feature generated from BERT~\cite{devlin2018bert} to compute the attention distribution conditioned on the previous hidden state of the \BaseLSTM: 
\begin{equation}
\begin{aligned}
    \alpha_t &= \text{softmax}( (W_x h_{t-1})^\top) x_t \\
    \hat{x_t} &= \alpha_t^\top x_t 
\end{aligned}
\end{equation}
where $W_x$ is the learnable parameter, $h_{t-1}$ is the previous \BaseLSTM hidden state, $x_t$ is the current language feature, and $\hat{x_t}$ is the weighted sum of $x_t$ over the attention distribution $\alpha_t$. Furthermore, each visual observation of the agent's view is encoded with a pre-trained Mask R-CNN~\cite{he2017mask}, where we take the scene feature from its ResNet-50-FPN backbone. At each time step $t$, the \BaseLSTM takes in the object feature $\{o_{t,z}\}$, which are 20 highest-scoring object features from the vision encoder concatenated with spatial and reachability encoding as in \BaseVLNBERT, scene feature $v_t$, language feature $\hat{x_t}$, previous action embedding $a_{t-1}$, and outputs a new hidden state $h_t$:
\begin{equation}
    h_t = \text{\BaseLSTM}([\{o_{t,z}\}; v_t;\hat{x_t};a_{t-1}], h_{t-1})
\end{equation}

The agent interacts with the environment by choosing an action and providing a binary mask (if the action is an interaction). To leverage the power of the pre-trained vision encoder, we follow \cite{pashevich2021episodic, singh2021factorizing} and ask our agent to choose an object $o_{t,z}$. The corresponding pixel mask is retrieved from the predicted object. We formulate object choosing in a same fashion as choosing navigable directions in common VLN models~\cite{fried2018speaker, tan2019learning, wang2021structured}. Action and object are generated from two different networks:
\begin{equation}
\begin{aligned}
    a_t &= \argmax(W_a[h_t;u_t]) \\
    p(o_{t,z}) &= \text{softmax}_z(o_{t,z}W_oh_t) \\
    \hat{o}_{t,z} &= \argmax_z p(o_{t,z})
\end{aligned}
\end{equation}

where $W_o$ and $W_a$ are learnable parameters, and $u_t = [v_t;\hat{x_t};a_{t-1}]$. Action prediction is trained with the ground-truth expert actions and reinforcement learning. The object feature is learned end-to-end with the ground-truth object information.

\subsection{Vision Encoder}
\label{s:vision}
For the vision encoder for \Milestone, we train an instance segmentation model, Mask R-CNN\cite{he2017mask}, as our vision encoder with training data generated from the expert demonstration images from ALFRED and ground-truth segmentation information from Ai2Thor simulator. Mask R-CNN is a two-stage detector, which the first stage proposes region of interests (RoI) by Region Proposal Network (RPN) \cite{ren2015faster} and the second stage extracts RoI features from the feature map by RoI Align \cite{he2017mask} and makes predictions with three heads, box classification, box regression, and mask head. The box heads share the same RoI features extracted with proposals, while the mask head extracts with the predictions by the box regression head. 

In the milestone checking, we not only are interested in objects and their locations but also care about whether they are reachable by the agent. Following the idea, we further implement the fourth head, using the same manner as the mask head, to predict availability for each object. We simply define objects within 1.5 meters as available for the binary classification by using the distance information from Ai2Thor simulator. Finally, the overall loss of pre-training our vision encoder is the summation of losses from four heads, 
\begin{equation}
    \mathcal{L} = \mathcal{L}_{cls} + \mathcal{L}_{reg} + \mathcal{L}_{mask} + \mathcal{L}_{avail}.
\end{equation}

With a pre-trained vision encoder, we use its ResNet \cite{he2016deep} backbone to encode scene features from environments and top-k RoI features from the mask head as object features to attend with the language model. For milestone checking, it also provides object labels and their reachability information. 

\begin{table*}
    \centering
    \captionsetup{width=.95\textwidth}
    \begin{tabular}{llcccc}
        \toprule
        \textbf{Error Types} & & \textbf{L} & \textbf{L+\milestone} & \textbf{V} & \textbf{V+\milestone}\\
        \midrule
        No error          &                  &     76      &  129 &  83  &  134 \\
        \midrule
        Interaction Failure &  & 231 &  255 & 178 &  199 \\
                            &    Collision &   87  & 99  & 100  &        121       \\
                            &   Interact with other object & 31  &   23  & 31  &   54\\
                            &  Wander endlessly &  66   &    130  & 22   &    24           \\
                            &  Navigate to next subtask location &  47 & 0 & 33 & 0 \\
        \midrule
        Navigation Failure  &    &     513 &    436 & 559 &    487 \\
                             &    Collision &   205  &   288 & 245  &   296       \\
                            &   Interact with other object    & 24  &  18  & 33  &  13 \\
                            &  Wander endlessly &   183  &  130 & 203  &  172 \\
                            &    Navigate to next subtask location &  101  & 0 & 78  & 0 \\
        \bottomrule
    \end{tabular}
\caption{\textbf{Error cases on unseen validation.} \textbf{Interaction Failure:} Agent gets close to the target object but fails to interact with it. \textbf{Navigation Failure:} Agent does not navigate to the target object at all. \textbf{Next Action:} Next action that happens after the failure has happened. \textbf{L} stands for \EnhancedLSTM and \textbf{V} stands for \EnhancedVLNBERT. }
\vskip -10pt
\label{tab:first_fatal}
\end{table*}

\subsection{Learning}
\subsubsection{Reward Shaping}
In addition to the progress (navigation) and stop rewards defined in \BaseVLNBERT~\cite{Hong_2021_CVPR}, we apply the interaction action matching as an additional reward to guide the agent to perform interaction action when needed. Furthermore, we introduce a visibility reward to make the agent learn to face the correct direction during the interaction.

\mypara{Navigation Reward}. Following \BaseVLNBERT, navigation reward acts as a strong supervision for directing our agent to the target object. Formally, $D_t$ is a distance from the agent to the target object at time $t$, and $\Delta D_t =  D_{t-1} - D_t$ is a change of distance by an action $a_t$. Reward for each $a_t$ is defined as: 

\begin{equation}
    r^D_t = \begin{cases}
    +1,& \Delta D_t > 0\\
    -1,              & \text{otherwise}
\end{cases}
\end{equation}

\mypara{Stop Reward.} When the agent decides to stop $a_t == \text{\fl{stop}}$, we give the agent the final reward depending on if the task is successful or not:

\begin{equation}
    r_{final} = \begin{cases}
    +3,& \text{Task} == \text{Success} \\
    -3,              & \text{otherwise}
\end{cases}
\end{equation}
where the task success means that the agent has completed all subtasks for the task.

\mypara{Interaction Reward}.
Since \BaseVLNBERT trains the agent in a navigation-only dataset, we need to define an additional interaction reward in order for the agent to learn to choose between navigation and interaction actions. Formally, at a given time step $t$, the agent will be rewarded if the interaction action matches the ground-truth interaction action, which we retrieve from the environment state. The reward for each $a_t$ is defined as:

\begin{equation}
    r^I_t = \begin{cases}
    +1,& a_t == a^*_t\\
    -1,              & \text{otherwise}
\end{cases}
\end{equation}

\mypara{Visibility Reward}. We define an additional visibility reward to ensure the agent learns to face the correct direction of the object to be interacted with before predicting an interaction action to that object. This reward is paired with the interaction reward, so the total reward that an agent can get from interacting with the right object is $+2$.

\begin{equation}
    r^V_t = \begin{cases}
    +1,& o^*_{t} \text{ is reachable}\\
    -1,              & \text{otherwise}
\end{cases}
\end{equation}

\subsubsection{Behavior Cloning}
Following the prevalent approach in training VLN models~\cite{Hong_2021_CVPR, tan2019learning}, we combine reinforcement learning and imitation learning (\ie, behavior cloning) to train our model. At each time step, the model is expected to produce the ground-truth action and interaction mask (for interaction actions). We apply cross-entropy loss between the predicted actions and the ground-truth action and add it to our reinforcement learning loss to ensure that the current trajectory is favored toward the expert demonstration trajectory. While we can adapt the expert demonstration directly on the \LSTMMilestone, it is not straightforward in \VLNBERTMilestone because it requires a panoramic input. Since ALFRED does not provide panoramic expert demonstration, we generated panoramic ground-truth trajectory information from ALFRED expert demonstration using its trajectory augmentation tool\footnote{https://github.com/askforalfred/alfred/tree/master/gen}.

\subsection{Training Details}

For the \BaseLSTM, we use a pre-trained BERT as the language encoder and randomly initialize the rest of the model. For \BaseVLNBERT, we use its pre-trained weights on R2R~\cite{anderson2018vision} to initialize the model. For the vision encoder in the milestone builder, we use a ResNet-50-FPN \cite{he2017mask} as the backbone for Mask R-CNN and finetune on ALFRED expert demonstrations from the Ai2Thor simulator \cite{ai2thor} with batch size 16. We finetune the Mask R-CNN pretrained on ImageNet \cite{ILSVRC15} on 4 Nvidia A6000 GPUs for 270k iterations, with a learning rate of 0.02, which is decreased by 10 at the 210k and 250k iteration. A weight decay of 0.0001 and momentum of 0.9 are applied. For all the experiments with \milestone and baseline models, we use a single Nvidia 2080TI GPU and AdamW optimizer is applied with a fixed learning rate of $10^{-5}$ for \BaseVLNBERT and $10^{-4}$ for \BaseLSTM. The batch size is set to 4, and the agent is trained for 20 epochs maximum. Early stopping is applied when the model shows no improvement on 3 consecutive epochs, and the model that shows the highest SR on the validation unseen split is adopted for testing. For all model training, only training split was used for training and validation split was held out.

\section{Additional Experiments}
\label{s:exp}

\begin{table}
  \centering
  \begin{tabular}{lcc}
        \toprule
        \textbf{Task Type}   & \textbf{Valid Unseen} & \textbf{Valid Seen}   \\
        \midrule
        Pick \& Place       & 48 & 63 \\
        Stack \& Place      & 28  & 25  \\
        Place Two           & 32 & 38 \\
        Examine             & 50 & 69\\
        Heat \& Place       & 19 &  18\\
        Cool \& Place       & 32 &  29  \\
        Clean \& Place      & 27 & 29  \\
        \bottomrule
  \end{tabular}
\caption{\textbf{Validation completion rate (\%) by task type for \VLNBERTMilestone.}}
  \label{tab:milestone_count}
\end{table}

\mypara{Agent frequently skips subtask}
As mentioned in the main paper, we perform an experiment that shows that the agent frequently skips a subtask and tries to execute the next subtask. In \autoref{tab:first_fatal}, we perform an analysis of the first fatal error that the agent encounters for the validation unseen split. We first categorize the failure type by \textit{interaction failure}, which the agent gets close to the interaction target but fails to perform an interaction action, and \textit{navigation failure}, where the agent does not get close to a target object at all. Then, we also retrieve the next sequence of actions after the failure and sort them into four categories:
\begin{itemize}
    \item \textbf{Collision}: Agent gets stuck in the environment and can not get out
    \item \textbf{Interaction with other object}: Agent interact with a wrong object or interacts when interaction is not needed
    \item \textbf{Wander endlessly}: Agent endlessly repeats a certain sequence of actions 
    \item \textbf{Navigate to the next subtask location}: Agent navigates to the location of the next target object in the next subtask
\end{itemize}
We show that \milestone performs better on most error cases in both interaction and navigation. Specifically, \milestone notably prevents the agent from performing the next subtask without completing the current one (\eg, \num{101} vs. \num{0}), reflecting \milestone's idea.

\mypara{\milestone completion rate by task type} We further analyze \milestone's completion rate by task type in \autoref{tab:milestone_count}. We can see that \milestone excels at completing relatively simple tasks such as \nl{Pick \& Place} or \nl{Examine} effectively, while suffers in complex tasks that require multiple interactions (\eg, \nl{Heat \& Place}). We attribute this phenomenon due to the fact that the current \milestone does not effectively learn the multi-interaction reasoning. However, by seeing the notable improvement on the simple tasks, \milestone opens up the possibility of leveraging more fine-grained milestone construction to improve the agent's task learning in VLN.

\end{document}